\documentclass{article}

\usepackage[preprint]{neurips_2025}

\usepackage{natbib}
\setcitestyle{numbers,square}

\usepackage[utf8]{inputenc} 
\usepackage[T1]{fontenc}    
\usepackage{hyperref}       
\usepackage{url}            
\usepackage{booktabs}       
\usepackage{amsfonts}       
\usepackage{nicefrac}       
\usepackage{microtype}      
\usepackage{xcolor}         

\usepackage{graphicx}
\usepackage{amsmath}
\usepackage{algorithm}
\usepackage{algorithmic}

\usepackage{graphicx}
\usepackage{wrapfig}

\usepackage{amsmath}

\usepackage{multirow,hhline}
\usepackage{subcaption}
\usepackage{colortbl}
\usepackage[capitalize,noabbrev]{cleveref}
\definecolor{hight_light}{RGB}{224, 241, 239}
\definecolor{darkgray}{gray}{0.4}

\usepackage{natbib}

\title{
C²Prompt: Class-aware Client Knowledge Interaction for Federated Continual Learning
}

\author{%
  Kunlun Xu\thanks{Equal contribution}  \\
  Wangxuan Institute of Computer Technology\\
   Peking University\\
  Beijing, China \\
  \texttt{xkl@stu.pku.edu.cn} \\
  \And
  Yibo Feng\textsuperscript{\rm *} \\
   Wangxuan Institute of Computer Technology\\
   Peking University\\
  Beijing, China \\
  \texttt{2022090917012@std.uestc.edu.cn} \\
  \AND
  Jiangmeng Li \\
  University of Chinese Academy of Sciences \\
   Beijing, China \\
  \texttt{jiangmeng2019@iscas.ac.cn} \\
  \And
  Yongsheng Qi \\
  Inner Mongolia University of Technology \\
  Hohhot, Inner Mongolia Autonomous Region \\
  \texttt{qys@imut.edu.cn} \\
  \And
  Jiahuan Zhou \thanks{Corresponding author} \\
  Wangxuan Institute of Computer Technology\\
   Peking University\\
   Beijing, China \\
  \texttt{jiahuanzhou@pku.edu.cn} \\
}

\begin{document}


\maketitle

\begin{abstract}

Federated continual learning (FCL) tackles scenarios of learning from continuously emerging task data across distributed clients, where the key challenge lies in addressing both temporal forgetting over time and spatial forgetting simultaneously. Recently, prompt-based FCL methods have shown advanced performance through task-wise prompt communication.
In this study, we underscore that the existing prompt-based FCL methods are prone to class-wise knowledge coherence between prompts across clients. The class-wise knowledge coherence includes two aspects: (1) intra-class distribution gap across clients, which degrades the learned semantics across prompts, (2) inter-prompt class-wise relevance, which highlights cross-class knowledge confusion. During prompt communication, insufficient class-wise coherence exacerbates knowledge conflicts among new prompts and induces interference with old prompts, intensifying both spatial and temporal forgetting.
To address these issues, we propose a novel \textbf{C}lass-aware \textbf{C}lient Knowledge Interaction (\textbf{C${}^2$Prompt}) method that explicitly enhances class-wise knowledge coherence during prompt communication. Specifically, a local class distribution compensation mechanism (LCDC) is introduced to reduce intra-class distribution disparities across clients, thereby reinforcing intra-class knowledge consistency. Additionally, a class-aware prompt aggregation scheme (CPA) is designed to alleviate inter-class knowledge confusion by selectively strengthening class-relevant knowledge aggregation. Extensive experiments on multiple FCL benchmarks demonstrate that C${}^2$Prompt achieves state-of-the-art performance. Our source code is available at \href{https://github.com/zhoujiahuan1991/NeurIPS2025-C2Prompt}{https://github.com/zhoujiahuan1991/NeurIPS2025-C2Prompt}
\end{abstract}

\section{Introduction}
With the proliferation of edge computing and IoT devices~\cite{zhang2021federated}, federated continual learning (FCL) has emerged as a critical paradigm for enabling intelligent systems to continuously learn from decentralized data streams while preserving data privacy~\cite{gao2024fedprok,li2024personalized,tran2024text,khan2024hydra}. However, this setting presents a dual challenge: models must overcome catastrophic forgetting across sequential tasks (temporal dimension) while adapting to heterogeneous data distributions among clients (spatial dimension)~\cite{li2024towards,dong2022federated}. While traditional continual learning methods~\cite{wang2022dualprompt,li2024fcs,xu2025dask,smith2023coda,zhou2025distribution} and federated learning approaches~\cite{li2024resource, huang2022learn, shi2024clip, wangtaming,morafahtowards} have made significant progress independently, their combined formulation in FCL struggles to address the superimposed forgetting effectively~\cite{li2024towards,babakniya2023data,piao2024federated}.

\begin{wrapfigure}{!t}{0.6\textwidth}
\includegraphics[width=0.6\textwidth]{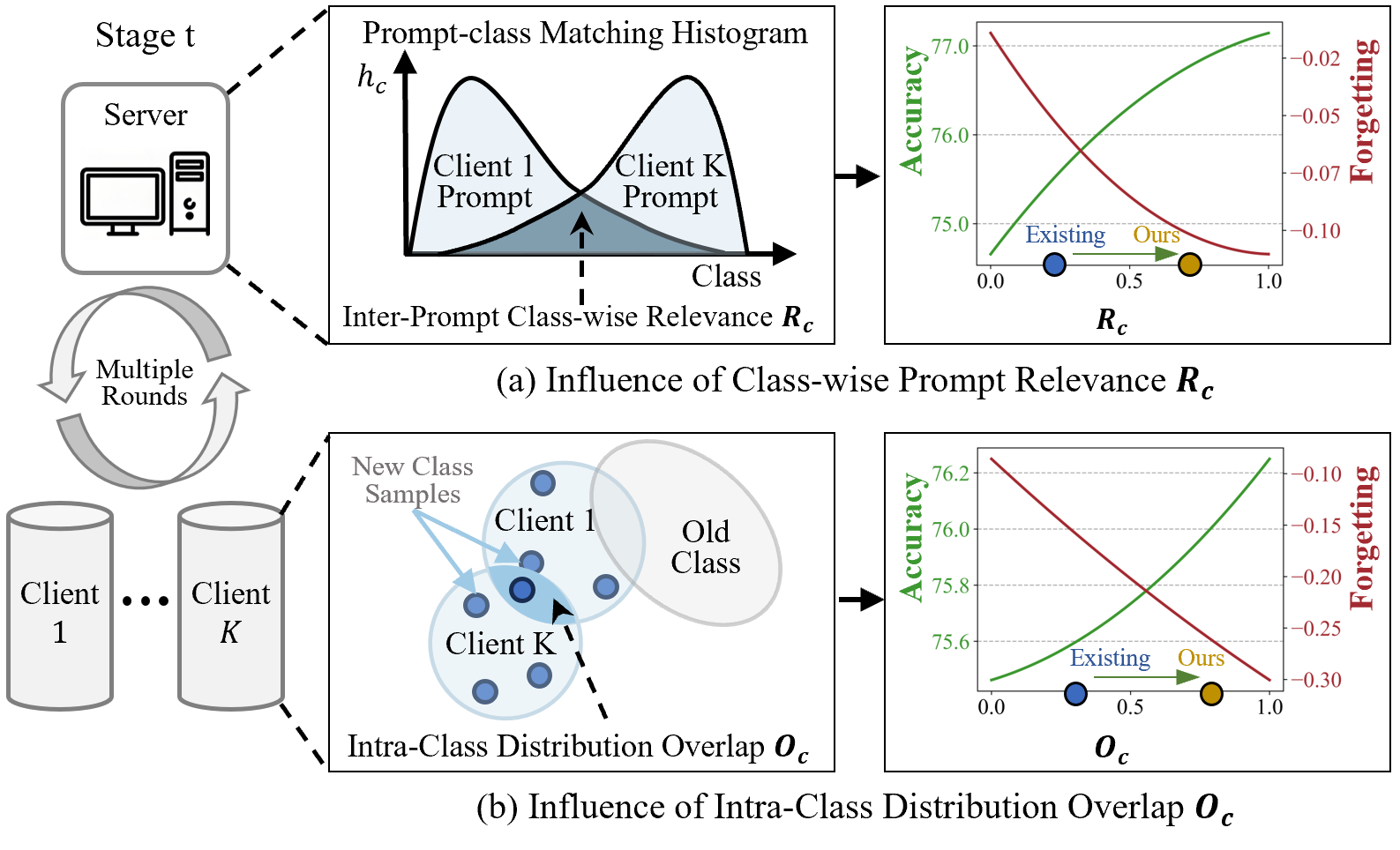}
\caption{In FCL, class-wise knowledge coherence includes two aspects: (a) inter-prompt class-wise relevance which influences prompt aggregation in the server, (b) intra-class distribution gap (overlap) across clients which influences the locally learned semantics of each class. }
\label{fig:first}
\end{wrapfigure}

Existing FCL methods predominantly address the challenges of spatio-temporal knowledge transfer through data synthesis and parameter regularization~\cite{NEURIPS2019_15825aee}. However, data synthesis approaches~\cite{ NEURIPS2019_e562cd9c, 9577808} typically depend on deep generative models trained on raw data, raising concerns regarding data privacy. In contrast, parameter regularization methods~\cite{li2024personalized} attempt to balance learning and forgetting but often suffer from limited capacity to acquire new knowledge effectively. Recently, prompt-based learning~\cite{piao2024federated,xu2025componential,ma2025federated} has emerged as a promising solution for FCL by maintaining task-specific prompts that store knowledge representations while leveraging a frozen pre-trained model~\cite{zhang2025scap,liu2025stop}. To overcome the overfitting to local distribution, some methods introduce inter-client prompt communication to improve the robustness~\cite{piao2024federated}. Despite their potential, these approaches are prone to class-wise knowledge coherence during prompt communication, which comprises two aspects.

First, as illustrated in Figure~\ref{fig:first} (a), class-wise knowledge across prompts from different clients inherently varies to some extent. During prompt communication (aggregation), this divergence often results in knowledge conflicts, degrading the model's acquisition capacity. Moreover, these conflicts exacerbate forgetting, as the aggregated prompts may conflict with the historical prompts. Second, as depicted in Figure~\ref{fig:first} (b), the intra-class distribution disparity across clients often affects the learned semantics of prompts. Certain features, although locally discriminative, may prove suboptimal from a global perspective, leading to further knowledge conflicts during prompt communication. Additionally, these locally discriminative features may be confused with historical data representations, resulting in degraded performance on previously learned tasks.

To address these challenges, we propose a novel Class-aware Client Knowledge Interaction (C${}^2$Prompt) approach to improve inter-prompt class-wise relevance and intra-class distribution overlap simultaneously, as shown in Figure~\ref{fig:first} (a)-(b). To achieve this, we first collect the local class distributions across clients and estimate the class-wise global distribution according to probability theory. Then, a local class distribution compensation mechanism (LCDC) is developed, which learns a set of class prompts to transfer the local semantics to the global domain, significantly improving intra-class knowledge consistency across clients.
Additionally, each local prompt is recorded with its affinity with different classes. Then, a class-aware prompt aggregation scheme (CPA) is designed to exploit the class affinities to estimate the class-wise knowledge relevance across prompts and generate dynamic weights to enhance class-relevant knowledge aggregation, effectively alleviating the confusion caused by class knowledge conflict.  Extensive experiments on multiple FCL benchmarks demonstrate that C${}^2$Prompt outperforms state-of-the-art methods by large margins.

 To summarize, the contributions of our paper are three-fold: 
 (1) We present the C${}^2$Prompt, an exemplar-free method that achieves Class-aware Client Knowledge Interaction to mitigate the temporal and spatial forgetting simultaneously. (2) A local class distribution compensation mechanism is developed to complement local distribution to improve cross-client semantic consistency. (3) A class-aware prompt aggregation scheme is proposed to enhance intra-class knowledge aggregation and alleviate the knowledge conflict via a class-wise knowledge relevant estimation mechanism. (4) The superiority of C${}^2$Prompt is validated on the challenging FCL benchmarks, where our method consistently achieves remarkable state-of-the-art performance.

\section{Related Work}
In this section, we review three research directions and discuss the state-of-the-art works that are most relevant to this paper. 
\subsection{Federated Learning}
FL considers a distributed machine learning paradigm where decentralized data resources are modeled collaboratively~\cite{liao2024foogd,chenclassifier,allouah2024fine,qi2025cross}. Each client trains with its corresponding data locally, and a server aggregates the client knowledge to obtain a global model~\cite{qi2023cross,pan2024federated,weng2024probabilistic,wufiarse}. The key challenge in FL is the data heterogeneity problem, where the data are not independently and identically distributed (non-IID) on different clients~\cite{zhang2024improving,mclaughlin2024personalized,m2024personalized,li2024global}. 
Current FL approaches can be primarily categorized into three branches, \textit{i.e.}, client-side regularization, server-side regularization, and synthetic data generation~\cite{zhang2024fedgmkd}. 
Client-side regularization methods aim to improve the alignment with the global model by refining local updates~\cite{li2020federated,karimireddy2020scaffold,mendieta2022local,acar2021federated,gao2022feddc,li2021model,lin2020ensemble,meng2024improving}. Server-side regularization approaches focus on achieving better aggregation to maximize the performance of the global model~\cite{mcmahan2016federated,hsu2019measuring,wang2020tackling,li2021fedbn}. 
Synthetic data generation methods rely on MixUp or training a deep generation model to generate synthetic data to approximate IID conditions~\cite{zhu2021data,guo2023pfedprompt} or post-train the global model~\cite{yoonfedmix,hu2022fedsynth,xiong2023feddm}. However, these FL methods assume that all the training data are available at the same time and neglect the practical condition that the training data occur sequentially.

\subsection{Continual Learning}
Continual Learning (CL) aims to learn with non-stationary data and generate a unified model that can address multiple tasks~\cite{li2024fcs,xu2024distribution,xu2025long}. 
Current CL methods are mainly divided into two categories: rehearsal-based and rehearsal-free. Rehearsal-based methods~\cite{NEURIPS2019_15825aee, NEURIPS2019_e562cd9c, 9577808} save a subset of learned samples into a memory buffer and replay them when learning a new task.  While promising performance has been achieved,  they usually require a large memory cost and raise privacy concerns during long-term learning. 
Rehearsal-free methods dynamically expand the network or isolate parameters for different tasks, regularize the network parameters that are important to learned tasks. Recently, freezing the pre-trained backbone model and only training a subset of learnable parameters is the current mainstream approach~\cite{liu2024compositional,wang2024hierarchical,xu2024mitigate,wang2022dualprompt}. 
 L2P~\cite{wang2022learning} pioneeringly introduced prompt learning to CL and proposed a key-query similarity method to select prompts for each task data from a prompt pool. CODAPrompt~\cite{smith2023coda} transforms prompt selection into a differential process with an attention mechanism. However, these approaches only consider alleviating temporal forgetting and struggle to address the non-IID data in the federated learning scenario~\cite{bai2024diprompt,xu2024lstkc,feng2023learning}.
 
\subsection{Federated Continual Learning}
In FCL, each client continuously learns from a private and incremental task stream locally and a global model aims to aggregate the spatial-temporal knowledge in a unified model~\cite{li2024personalized}.  Existing FCL methods primarily focus on generative replay to address the spatial and temporal forgetting~\cite{liang2024diffusion,babakniya2023data, li2024towards,qi2023better,zhang2023target}. However, due to the slow convergence of generation training, training a generative model introduces massive training overheads~\cite{odena2017conditional}. Besides, the generative models typically risk privacy leakage of local information~\cite{li2024towards}. Recently, efficient tuning-based methods have shown advanced performance in FCL. PILoRA and LoRM introduced LoRA to address FCL by learning low-rank parameters in each client and aggregates them in the server. Besides, prompt learning has shown remarkable anti-forgetting capacity due to the parameter-matching mechanism that enables mutli-task knowledge co-consistency~\cite{yu2024personalized,bagwe2023fedcprompt,piao2024federated}. However, existing methods typically neglect the knowledge conflict between individual prompts during server-side aggregation, leading to significant knowledge loss.

\begin{figure}[t]
    \centering
	\includegraphics[width=1.0\linewidth]{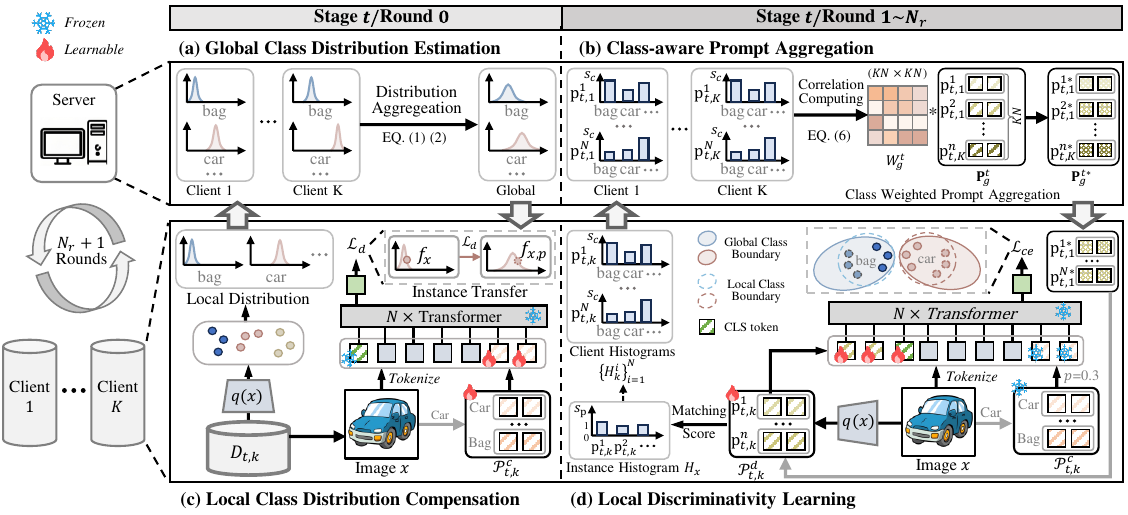} 
        \caption{\label{fig:framework} Overview of our C${}^2$Prompt approach. During the training stage $t$, given the data $D_{t,k}$ at each client $k$, the local class-aware feature distribution is collected and uploaded to the server to estimate the global distribution of each class. Then, the global distribution is distributed to the local clients to train client-specific class-distribution compensation prompts $\mathcal{P}_{t,k}^c$. During the later process, the discriminativity prompts $\mathcal{P}_{t,k}^d$ are introduced to learn classification-relevant knowledge, which are iteratively aggregated in the server according to the class knowledge relevance for $N_r$ rounds. }      
\end{figure}

\section{Preliminaries}
\textbf{Problem Formulation:}
In FCL, a collection of $K$ clients collaboratively learn under the coordination of a central server. Each client $k \in \{1, 2, \dots, K\}$ sequentially learns a series of $T$ tasks. Let $\mathcal{T}_k^{t}$ denote the $t$-th task of the $k$-th client, and $D_k^{t}$ be its corresponding dataset. The model parameters of client $k$ during the learning of $\mathcal{T}_k^{t}$ are represented as $\theta_k^t$. 

\textbf{Baseline:} 
Following recent FCL methods~\cite{piao2024federated,halbe2023hepco,bagwe2023fedcprompt,piao2024federated}, we adopt CODAPrompt~\cite{smith2023coda} as the basic architecture for both clients and the server. For each local task $\mathcal{T}_k^{t}$, a set of prompts $\mathbf{P}_k^{t} \in \mathbb{R}^{N \times L_p \times D}$ is learned, where $N$ is the number of local prompts, $L_p$ is the length of each prompt, and $D$ is the input dimension of the Vision Transformer (ViT) encoder.
On the server side, a global prompt pool $\mathbf{P}_g \in \mathbb{R}^{N_g \times L_p \times D}$ is maintained, containing prompts from both previous tasks $\mathcal{T}^{pre}$ and the current task $\mathcal{T}^{cur}$. The total number of prompts in the pool is denoted as $N_g = M \times N$, where $M$ is the number of seen tasks.
For an image $\boldsymbol{x}$, its associated prompt $\mathbf{p}_x \in \mathbb{R}^{L_g \times D}$ is generated through a weighted sum of the prompts in $\mathbf{P}_g$:

\begin{equation}  
\mathbf{p}_x=\sum_i^{N_g}\alpha_i[\mathbf{P}_g]_i\:,
 \label{eq:coda}
\end{equation}

where the weights $\boldsymbol{\alpha}_{\boldsymbol{x}}=\{\alpha_1,\alpha_2,\ldots,\alpha_{M_g}\}$ are computed based on query-key similarity:
\begin{equation} 
\boldsymbol{\alpha}_{\boldsymbol{x}}=\{\gamma(q(x)\odot[\mathbf{A}_g]_1,[\mathbf{K}_g]_1),\gamma(q(x)\odot[\mathbf{A}_g]_2,[\mathbf{K}_g]_2),
\ldots,\gamma(q(x)\odot[\mathbf{A}_g]_{N_g},[\mathbf{K}_g]_{N_g})\},
\end{equation}
where $\gamma(\cdot,\cdot)$ represents the cosine similarity function, $\mathbf{K}_g\in\mathbb{R}^{M_g\times D}$ and $\mathbf{A}_g\in\mathbb{R}^{M_g\times D}$ are the learnable keys and attention weights of the prompts in $\mathbf{P}_g$. $\odot$ denotes the Hadamard product. 
For simplicity, given the one-to-one correspondence among $\mathbf{A}_g, \mathbf{K}_g$, and $\mathbf{P}_g$, we represent the global prompt pool as $\mathbf{P}_g$, encapsulating both attention and key representations.

In addition to the basic architecture of CODAPrompt, we also incorporate the knowledge distillation loss introduced by Powder~\cite{piao2024federated} to enhance knowledge retention across tasks. The distillation loss is formulated as follows:

\begin{equation}
\mathcal{L}_{kd}(\hat{y}_{cu},\hat{y}_{tr})=-\sum_{k=0,k\neq y}^K[\hat{y}_{tr}]_k\mathrm{log}\frac{[\hat{y}_{cu}]_k}{[\hat{y}_{tr}]_k},
\end{equation}


where $\hat{y}_{cu}$ denotes the output logits of the current model, and $\hat{y}_{tr}$ represents the logits of the model at the beginning of the current communication round, containing the latest knowledge transferred from other tasks.

\section{Proposed Method}
In this section, we elaborate on our C${}^2$Prompt which primarily consist of four modules, \textit{i.e.}, Global Class Distribution Estimation, Class-aware Prompt Aggregation, Local Class Distribution Compensation, and Local Discriminativity Learning.  
An overview of C${}^2$Prompt is illustrated in Figure~\ref{fig:framework}, and the procedure is summarized in Algorithm~\ref{alg:framework} of Appendix~\ref{sec:alg}.

\subsection{Global Distribution Generation}
When the new stage data of a client $D_k^t$ is given,  the local distribution for each class is first computed, resulting in a distribution set $\mathcal{D}_k^t=\{(\mu_{k,i}^t,\sigma_{k,i}^t)\}_{i=1}^{|\mathcal{C}_k^t|}$, where $(\mu_{k,i}^t,\sigma_{k,i}^t)$ denotes the class center and standard deviation, and $|\mathcal{C}_k^t|$ is the number of classes for client $k$. For each class $i$, the data distribution on client $k$ is approximated by a Gaussian distribution $\mathcal{N}(\mu_{i,k}^t, (\sigma_{i,k}^t)^2)$. Furthermore, the proportion of samples for class $i$ at client $k$ relative to the global sample size of class $i$ is represented as $p_{k,i}^t$. The global mean and standard deviation are then aggregated across all clients as follows:
\begin{equation}        
\mu_i^g=\sum_{k=1}^K\mu_{i,k}^tp_{k,i}^t,
    \label{eq:global-mu}
\end{equation}
\begin{equation}        
(\sigma_i^g)^2=\sum_{k=1}^K\left((\mu_{i,k}^t)^2+(\sigma_{i,k}^t)^2\right)p_{k,i}^t-(\mu_i^g)^2
    \label{eq:global-sigma}
\end{equation}
The \textbf{theoretical derivations} of Equation~\ref{eq:global-mu} and Equation~\ref{eq:global-sigma} are provided in the Appendix~\ref{sec:Server-side}. Once calculated, the global distribution of each class is sent back to each client for further processing.
\subsection{Local Class Distribution Compensation}

Upon receiving the global class distribution, local class distribution compensation prompts $\mathcal{P}^{c}_{t,k}=\{\mathbf{p}_i^c\}_{i=1}^{|\mathcal{C}_k^t|}$ are introduced to address the issue of local undersampling by transferring the local samples to align with the global distribution. 
Specifically, for each class $i$, a local class distribution compensation prompt is denoted as $\mathbf{p}_i^c \in \mathbb{R}^{L_c \times d}$, where $L_c$ represents the length of $\mathbf{p}_i$.
Given an input image $\boldsymbol{x}$, it is first tokenized~\cite{li2024exemplar} into a sequence representation $\boldsymbol{h}_{\boldsymbol{x}} \in \mathbb{R}^{L_{\boldsymbol{h}} \times d}$, where $L_{\boldsymbol{h}}$ is the sequence length. The associated local class distribution compensation prompt of $\boldsymbol{x}$, denoted $\mathbf{p}_{\boldsymbol{x}}^c$, is obtained by indexing from $\mathcal{P}^{c}_{t,k}$ with its label. Both $\mathbf{p}_{\boldsymbol{x}}^c$ and $\boldsymbol{h}_{\boldsymbol{x}}$ are then fed into transformer layers:

\begin{equation}        f_{x,p}=\boldsymbol{f}_\theta([\boldsymbol{h}_{\boldsymbol{x}},\mathbf{p}_{\boldsymbol{x}}^c,cls]),
    \label{eq:cosine}
\end{equation}

where $\boldsymbol{f}_\theta$ is the parameters of the pretrained ViT, $cls$ is the [CLS] token~\cite{liu2024compositional}, and $f_{x,p} \in \mathbb{R}^c$ is the generated feature.
To ensure that $f_{x,p}$ aligns with the global class distribution, we assume that the global distribution for class $i$ follows a Gaussian parameterization $\mathcal{N}(\mu_i^g, (\sigma_i^g)^2)$. The alignment is enforced through a distribution-based cross-entropy loss that maximizes the likelihood of $f_{x,p}$ under the global distribution:
\begin{equation}       
\mathcal{L}_c=-\frac{1}{2}(f_{x,p}-\boldsymbol{\mu}_i^g)^\top(\boldsymbol{\Sigma}_i^g)^{-1}(f_{x,p}-\boldsymbol{\mu}_i^g)
    \label{eq:gaussian-loss},
\end{equation}
where $\boldsymbol{\Sigma}_i^g$ is a diagonal covariance matrix with its diagonal entries equal to $(\sigma_i^g)^2$. The \textbf{theoretical derivations} of Equation~\ref{eq:gaussian-loss} are provided in the Appendix~\ref{sec:Client-side}.
Note that once $\mathcal{P}^{c}_{t,k}=\{\mathbf{p}_i^c\}_{i=1}^{|\mathcal{C}_k^t|}$ is trained, it is frozen during the sequential rounds of training in the current stage.

\subsection{Local Discriminativity Learning}
When $\mathcal{P}^{c}_{t,k}$ is learned, we introduce the local discriminativity prompts $\mathcal{P}_{t,k}^d=\{\mathbf{p}_{t,k}^i\}_{i=1}^{N}$, matrixed as $\mathbf{P}_{t,k}^d$, which corresponds to prompts of original CODAPrompt, to enable new knowledge learning. Given $\boldsymbol{x}$ and its label $y$, we generate instance-specific discriminativity prompt $\mathbf{p}_{\boldsymbol{x}}^d$ form $\mathbf{P}_{t,k}^d$ according to Equation~\ref{eq:coda}. Besides, the local class distribution compensation prompt $\mathbf{p}_{\boldsymbol{x}}^c$ is indexed from $\mathcal{P}_{t,k}^c$ using $y$. Then, a cross entropy loss ($CE$) is introduced to optimize $\mathbf{p}_{\boldsymbol{x}}^d$:

\begin{equation}        \mathcal{L}_{ce}=CE\big(\mathbf{W}_k\boldsymbol{f}_\theta([\boldsymbol{h}_{\boldsymbol{x}},\mathbf{p}_{\boldsymbol{x}}^c,\mathbf{p}_{\boldsymbol{x}}^d,cls]),y\big),
    \label{eq:ce-loss}
\end{equation}

where $\mathbf{W}_k$ is the learnable weight of the classifier of client $k$. Note that $\mathbf{p}_{\boldsymbol{x}}^c$ is exploited with $p=0.5$ to sufficiently utilize the information of both local original data and the completed distributions.

At the same time, a instance histogram $H_{\boldsymbol{x}}=\{s_p^i\}_{i=1}^{N}$ for $\boldsymbol{x}$ is generated where $\{s_p^i\}$ denotes the similarity score between $\boldsymbol{x}$ and $\mathbf{p}^i_{t,k}$. During one round of training, we introduce a client histogram $H_k^i=\{s_c^j\}_{j=1}^{|\mathcal{C}_{k}^t|}$ for each prompt of stage $t$ that records the cumulative prompt-class matching scores $s_c^j$, which is mathematically calculated by: 
\begin{equation}        
s_c^j=\sum_{n=1}^{|D_{t,k}|}[H_{\boldsymbol{x}_n}]_j,
    \label{eq:client-histogram}
\end{equation}
where $s_c^j$ represents the affinity between the prompt and class.
Note that $H_k^i$ can be generated online during training and requires almost no additional computing overhead. When one round of discriminative prompt training is finished, a set of client histograms $\{H_k^i\}_{i=1}^N$ for the new stage prompts is uploaded to the server.

\subsection{Class-aware Prompt Aggregation}
When a round of local discriminativity learning is finished, the local client histograms are collected to form a set $\mathcal{H}_g^t=\{H_1^i\}_{i=1}^{N}\cup\{H_2^i\}_{i=1}^{N}\cup\cdots\cup\{H_K^i\}_{i=1}^{N}$, which is matrixed as $\mathbf{H}_g^t\in\mathbb{R}^{KN\times |\mathcal{C}_t|}$.
Then, an inter-prompt correlation matrix $W_g^t\in\mathbb{R}^{KN\times KN}$ is computed by
\begin{equation}
    W_g^t=\gamma(\mathbf{H}_g^t{\mathbf{H}_g^{t^\top}}/\tau),
    \label{eq:correlation}
\end{equation}
where $\gamma$ is the softmax function that is conducted row-wise here, and $\tau$ is a hyperparameter to scale the similarity scores. Besides, the prompts of stage $t$ are also collected from the clients to form a set $\mathcal{P}_g^t=\{p_1^i\}_{i=1}^{N}\cup\{p_2^i\}_{i=1}^{N}\cup\cdots\{p_K^i\}_{i=1}^{N}$, which is matrixed as $\mathbf{P}_g^t\in\mathbb{R}^{KN\times L_p\times d}$. Then, a Class Weighted Prompt Aggregation process is conducted by:

\begin{equation}
    \mathbf{P}_g^{t*}=W_g^t\mathbf{P}_g^t,
    \label{eq:weighted-aggregation}
\end{equation}

where $\mathbf{P}_g^{t*}\in\mathbb{R}^{KN\times L_p\times d}$ is the updated prompts that have collected the most relevant knowledge from prompts of different clients. Then, $\mathbf{P}_g^{t*}$ is split into $K$ prompt sets and distributed to the corresponding clients.

\textbf{Training and Inference:} 
As shown in Figure~\ref{fig:framework}, during stage $t$, the training process consists of two phases. Firstly, Global Class Distribution Estimation and Local Class Distribution Compensation are conducted at round 0. The local distribution compensation loss $\mathcal{L}_c$ is adopted to train $\mathcal{P}_{t,k}^c$. Then, from round 1 to $N_r$, Class-aware Prompt Aggregation and Local Discriminativity Learning are conducted in turn. The model is optimized by an overall loss:

\begin{equation}
    \mathcal{L}_d=\mathcal{L}_{ce}+\beta \mathcal{L}_{kd},
    \label{eq:total-loss}
\end{equation}

where $\beta$ is a hyperparameter to balance the loss components.

During inference, following previous works~\cite{piao2024federated}, the prompts learned on all the seen local tasks are collected to generate a prompt $\mathbf{p}_{\boldsymbol{x}}$ which is exploited to generate predictions by

\begin{equation}    
\hat{y}=\gamma\big(\mathbf{W}_g\boldsymbol{f}_\theta([\boldsymbol{h}_{\boldsymbol{x}},\boldsymbol{p}_{\boldsymbol{x}},cls]),y\big),
    \label{eq:prediction}
\end{equation}

where $\mathbf{W}_g$ is the global classifier by concentrates the local classifiers learned from different tasks following~\cite{piao2024federated}.

\section{Experiments}
\subsection{Experimental Setups}
\textbf{Datasets and Metrics:} We conduct the experiments on three widely used benchmarks in FCL, \textit{i.e.}, ImageNet-R\cite{hendrycks2021many}, DomainNet\cite{peng2019moment} and CIFAR-100\cite{krizhevsky2009learning}. To evaluate the effectiveness of different FCL methods, 6 metrics are adopted in this paper, including Average Accuracy (Avg), Average Incremental Accuracy (AIA), Forgetting Measure (FM), Forward Transfer (FT), Backward Transfer (BT), Combined Transfer (CT). The configurations of the benchmarks and the details of the metrics are presented in Appendix \ref{appendix:D}.


\begin{table*}[htbp]
  \centering
    \caption{\label{tab:all} Result comparison on the ImageNet-R and DomainNet benchmark}\label{table1} 
    \setlength{\tabcolsep}{0.1mm}{
    \begin{tabular*}{\textwidth}{@{\extracolsep{\fill}}clc>{\columncolor{hight_light}}c>{\columncolor{hight_light}}ccccc>{\columncolor{hight_light}}c>{\columncolor{hight_light}}ccccc}

        \toprule
        \multicolumn{2}{c}{\multirow{2}{*}{Methods}}&\multirow{2}{*}{Pub.}&\multicolumn{6}{c}{ImageNet-R}&\multicolumn{6}{c}{DomainNet}\\
        \cmidrule{4-15}
        &&& Avg$\uparrow$ & AIA$\uparrow$ & FM$\downarrow$ & FT$\uparrow$ & BT$\uparrow$ & CT$\uparrow$&Avg$\uparrow$ & AIA$\uparrow$ & FM$\downarrow$ & FT$\uparrow$ & BT$\uparrow$ & CT$\uparrow$\\
        \midrule
        &FedWEIT   &\scriptsize{\textcolor{darkgray}{\textit{ICML2021}}}&71.10&74.30&1.80&-2.39&-1.83&-3.86&67.84&69.63&1.91&-2.92&-3.11&-4.97  \\
        &CFeD    &\scriptsize{\textcolor{darkgray}{\textit{IJCAI2022}}}&47.93&59.79&3.81&-17.67&-14.92&-29.60&42.85&60.19&1.65&-4.98&-13.32&-15.64\\
        &GLFC       &\scriptsize{\textcolor{darkgray}{\textit{CVPR 2022}}}  & 72.96&75.21&1.10&-3.87&-1.55&-5.11&69.75&70.34&1.23&-4.08&-2.46&-6.04\\
        &Fedspace &\scriptsize{\textcolor{darkgray}{\textit{CVPR 2023}}}  &72.27&73.36&2.01&-2.60&-4.91&-5.95&68.98&70.71&1.80&1.87&-4.16&-1.45 \\
        \midrule
        &Fed-L2P& \scriptsize{\textcolor{darkgray}{\textit{CVPR2022}}}    &77.88  & 75.03  & 0.41  & -2.79  & -0.17 & -2.92 &70.98&72.36&0.16&-2.18&0.10&-2.09\\
        &Fed-Dual & \scriptsize{\textcolor{darkgray}{\textit{ECCV2022}}}          &76.85  & 74.91  & 0.49  & -3.12  & 0.22 & -2.95 &71.90&72.15&0.16&-1.82&\underline{0.41}&-1.49 \\
        &Fed-CODA&\scriptsize{\textcolor{darkgray}{\textit{CVPR2023}}}    &79.65 & 75.14 & \textbf{-0.68} & -2.53 & \underline{1.69} & -1.18 &72.47&72.84&\underline{0.01}&-0.82&\textbf{0.83}&-0.15\\
        &Fed-CP & \scriptsize{\textcolor{darkgray}{\textit{ICML2023}}} &76.75 & 72.59 & 0.63 & -3.16 & 0.00 & -3.16 &71.28&69.92&0.18&-2.78&0.00&-2.78 \\        
        &Powder &  \scriptsize{\textcolor{darkgray}{\textit{ICML2024}}} & \underline{84.69} & \underline{84.08} & \underline{-0.54} & \underline{4.48} & \textbf{1.95} & \underline{6.04}  &\underline{75.98}&\underline{77.28}&0.10&\underline{1.28}&0.14&\underline{1.40}    \\
        &PILoRA&  \scriptsize{\textcolor{darkgray}{\textit{ECCV2024}}}&45.43&48.72&0.92&-5.75&-7.32&-12.54&31.22&40.76&0.55&-0.12&-0.74&-1.81\\
        &Fed-MOS&  \scriptsize{\textcolor{darkgray}{\textit{AAAI2025}}}&47.67&47.08&1.40&-3.30&-0.12&-3.37&40.37&45.22&0.31&-1.43&-1.21&-2.50\\        
       & LoRM &  \scriptsize{\textcolor{darkgray}{\textit{ICLR2025}}}&58.00&67.78&8.71&-4.67&-9.22&-13.70&23.18&28.49&5.72&-1.32&-0.11&-1.40\\
        \midrule
        &C${}^2$Prompt&\scriptsize{\textcolor{darkgray}{\textit{This Paper}}}  &\textbf{87.20 }& \textbf{85.93} & -0.36 &\textbf{ 7.63}&1.12&\textbf{8.52} & \textbf{78.88}&\textbf{77.55}&\textbf{-0.02}&\textbf{3.87}&0.23&\textbf{4.05}\\

        \bottomrule
    \end{tabular*}}
\end{table*}%

\textbf{Compared Methods:} We compare our proposed C${}^2$Prompt with the following methods: (1) Fully-Tuning-based (FULLY) federated continual learning methods, including FedWEIT~\cite{yoon2021federated}, CFeD~\cite{ma2022continual}, GLFC~\cite{dong2022federated} and FedSpace~\cite{shenaj2023asynchronous}. (2) Efficient-Tuning-based (EFFICIENT) methods, including prompt learning approaches, FedCPrompt~\cite{bagwe2023fedcprompt} and Powder~\cite{piao2024federated}. Besides, the state-of-the-art prompt-based continual learning methods, \textit{i.e.}, L2P~\cite{wang2022learning}, DualPrompt~\cite{wang2022dualprompt}, CODAPrompt~\cite{smith2023coda}, are integrated with the well-known FedAvg~\cite{mcmahan2016federated} algorithm to make a comprehensive comparison (Fed-L2P, L2P-Dual, Fed-CODAP, Fed-CPrompt). Additionally, the LoRA-based FCL methods, including PILoRA~\cite{guo2024pilora} and LoRM~\cite{salami2024closed}, and the adapter-based continual learning method MOS~\cite{sun2025mos} is integrated with FedAvg to form Fed-MOS. All experiments are implemented using official code, with the ViT-B/16 pre-trained on ImageNet-21k serving as the backbone network.

\textbf{Implementation Details} 
The settings of our discriminativity prompts follow the configuration of previous works~\cite{piao2024federated}, where $L_p$, $N$ and $d$ are set to 10, 8 and 768, respectively. 
For our class distribution compensation prompt, the prompt length $L_c$ is set to 3 by default. The Adam optimizer with a learning rate of 0.01 is adopted during training. For all the experiments, the training and testing images are resized to 224×224. The client number $K$ and round number for each task are set to 5 and 3, respectively. All experiments are conducted on a single Nvidia 4090 GPU. 

\subsection{Comparison Results} 
We follow the experimental setting of the previous methods~\cite{piao2024federated} and the comparison results on the ImageNet-R and DomainNet are represented in Table~\ref{table1}, where Avg and AIA are the most important metrics indicating the long-term knowledge accumulation and progressive performance, respectively. The best and second best methods are highlighted in \textbf{Bold} and \underline{Underlined}, separately. 

\textbf{Avg Comparison:} Our C${}^2$Prompt outperforms the state-of-the-art Powder, achieving improvements of \textbf{2.51\%} and \textbf{2.90\%} on ImageNet-R and DomainNet, respectively. These results demonstrate the superiority of our method in long-term knowledge consolidation. This is because the new knowledge acquisition capacity is significantly improved with our local class distribution compensation and class-aware discriminativity prompt aggregation designs. Besides, the accurate knowledge communication mechanism avoids the irrelevant prompts fusion that generate invalid prompts which not only semantically away from new prompts, but also conflict with historical prompts.

\begin{wrapfigure}{!t}{0.55\textwidth}
\includegraphics[width=0.55\textwidth]{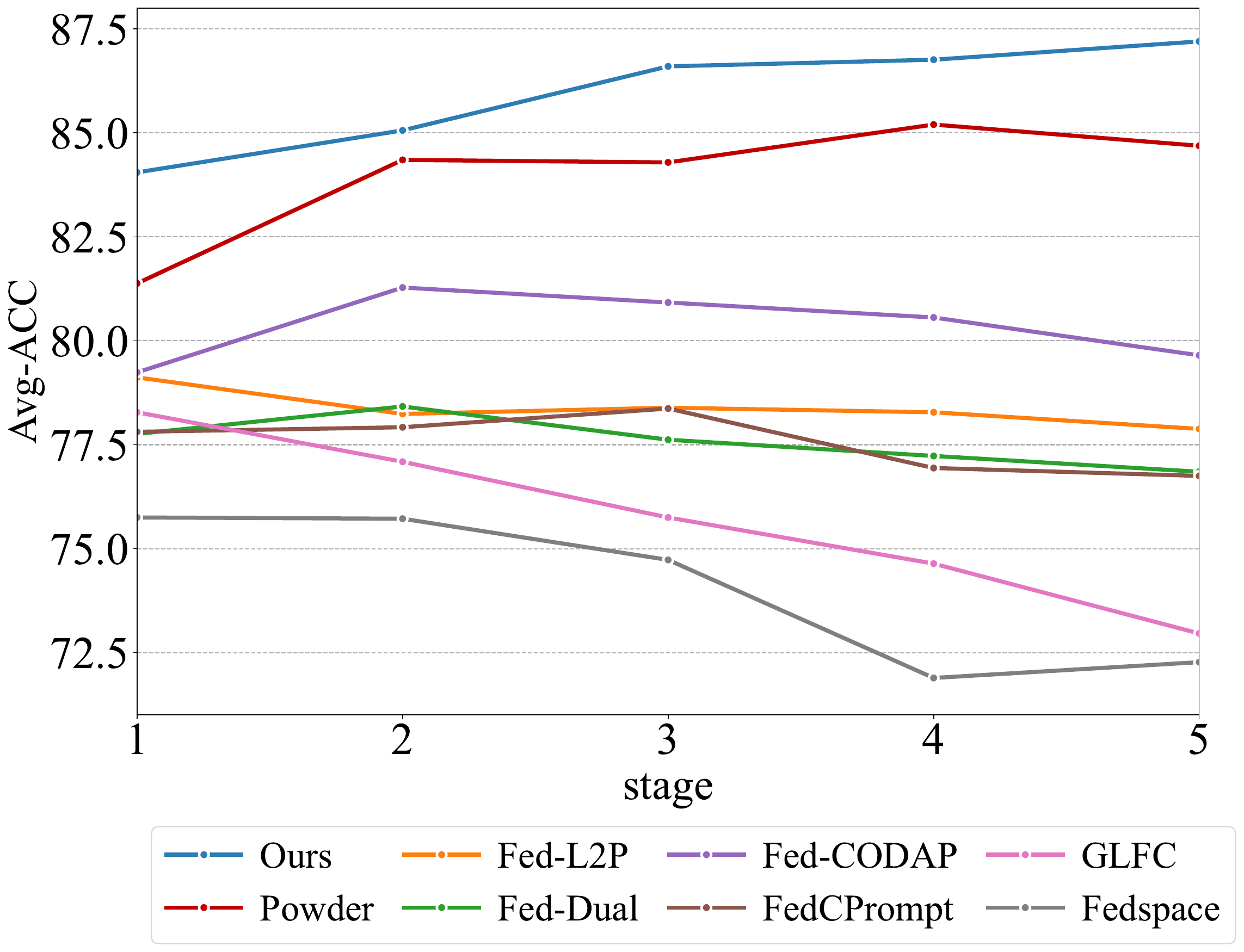}
\caption{Avg-ACC curves on the seen tasks across training stages .}
\label{fig:Avg-curve}
\end{wrapfigure}

\textbf{AIA Comparison:} Our C${}^2$Prompt achieves improvements of \textbf{1.85\%} on ImageNet-R and also outperforms all existing approaches on DomainNet, verifying our method consistently obtains superior performance compared the existing methods across different training stages. This is attributed to the local class distribution compensation and class-aware discriminativity prompt aggregation designs that enhance robust local knowledge acquisition and improve distributed knowledge collection.

\textbf{FM Comparison:} Fed-CODAP, Powder and our C${}^2$Prompt show a negative forgetting rate on the small-scale dataset ImageNet-R. This indicates the new tasks can facilitate historical task learning when training samples are limited. On the large-scale dataset DomainNet, only our C${}^2$Prompt shows a negative forgetting rate. These results verify the effective antiforgetting capacity of our method under different conditions. 

\textbf{FT Comparison:} Our C${}^2$Prompt shows advanced forward-transfer capacity, outperforming existing methods by at least \textbf{3.15\%} and \textbf{2.59\%} on ImageNet-R and DomainNet, respectively. This is primarily attributed to the Global Class Distribution Estimation and Local Class Distribution Compensation designs, where the former can effectively exploit the asynchronously arriving data of the same class to generate reliable global distribution estimation. Then the estimated global distribution is exploited by the later to achieve data-level information compensation, thereby significantly improving subsequent data learning.

\begin{wrapfigure}{!t}{0.55\textwidth}
\includegraphics[width=0.55\textwidth]{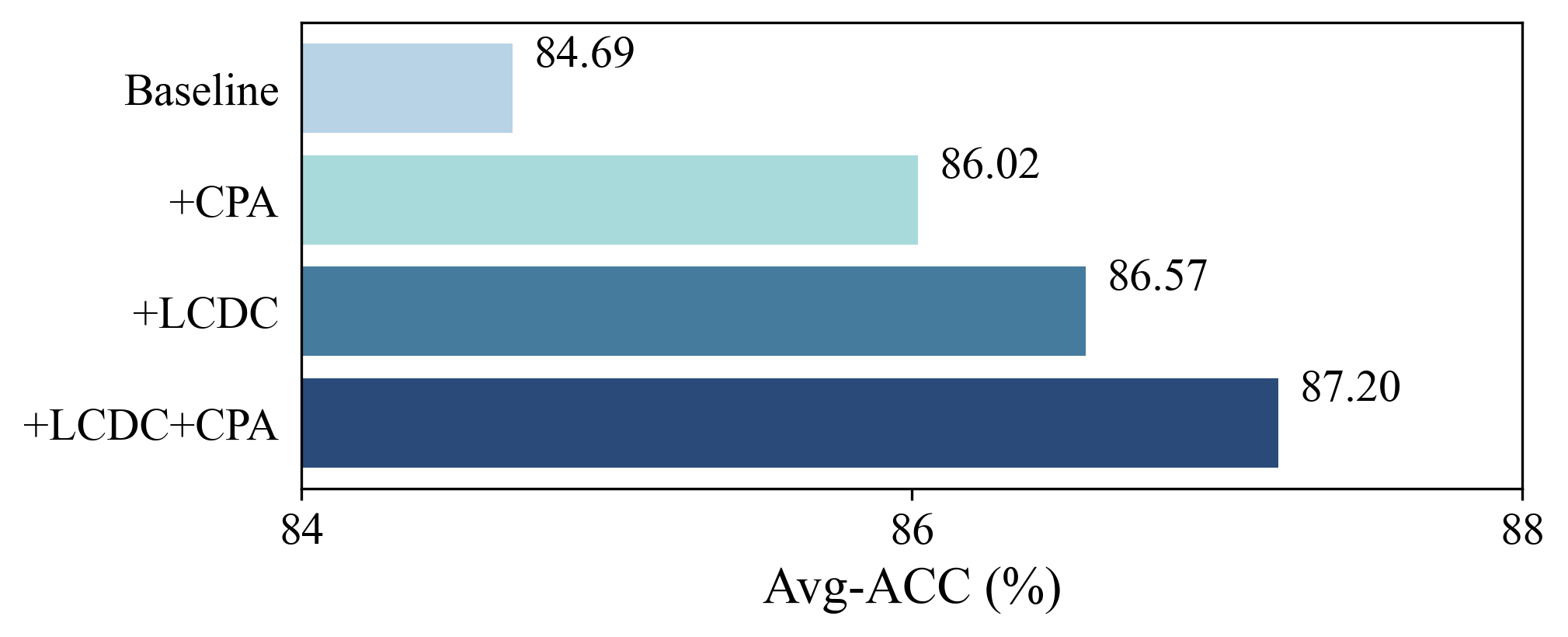}
\caption{Ablation on the model components.}
\label{fig:ablation}
\end{wrapfigure}

\textbf{BT Comparison:} Fed-CODAP, Powder and our C${}^2$Prompt consistently show positive backward-transfer across both ImageNet-R and DomainNet datasets. This is because the asynchronously arriving data enable the later tasks to enhance the knowledge of previously seen classes. We observe that the backward-transfer results of C${}^2$Prompt are relatively inferior to Fed-CODAP and Powder. This is because the Local Class Distribution Compensation and Class-aware Prompt Aggregation designs significantly improve the distributed data learning capacity at each stage, leaving less improvement space for seen tasks.

\textbf{CT Comparison:} As for the combined transfer of forward and backward, our C${}^2$Prompt outperforms all existing approaches with \textbf{2.48\%} and \textbf{2.65\%} improvements on ImageNet-R and DomainNet, respectively. These results demonstrate that the class-aware client knowledge interaction designs in this paper effectively boost the overall learning capacity in the temporal dimension in FCL. Specifically, the Global Class Distribution Estimation effectively aggregates the distributional information across spatial and temporal data sources. Local Class Distribution Compensation module leverages the global distributional image to overcome the non-IID phenomenon across clients. Finally, Local Discriminativity Learning and Class-aware Prompt Aggregation modules effectively integrate the distributional knowledge into the prompts.

\textbf{Performance Tendency Analysis:} To further analyze the model learning process, we visualize the Average Accuracy (Avg-ACC) across the seen tasks during the FCL stages in Fig.~\ref{fig:Avg-curve} on the ImageNet-R benchmark. The results show that our method consistently outperforms state-of-the-art approaches across all stages. Furthermore, the Avg-ACC of C${}^2$-Prompt exhibits a stable upward trajectory throughout the training process, whereas other methods display either declining or fluctuating trends. This advantage is attributed to our class-aware client knowledge interaction designs, which effectively extract and preserve robust knowledge over long-term training. In contrast, existing methods are more prone to knowledge conflicts during parameter aggregation in FCL, leading to performance degradation as training progresses.

\begin{figure}[t]
    \centering
	\includegraphics[width=1.0\linewidth]{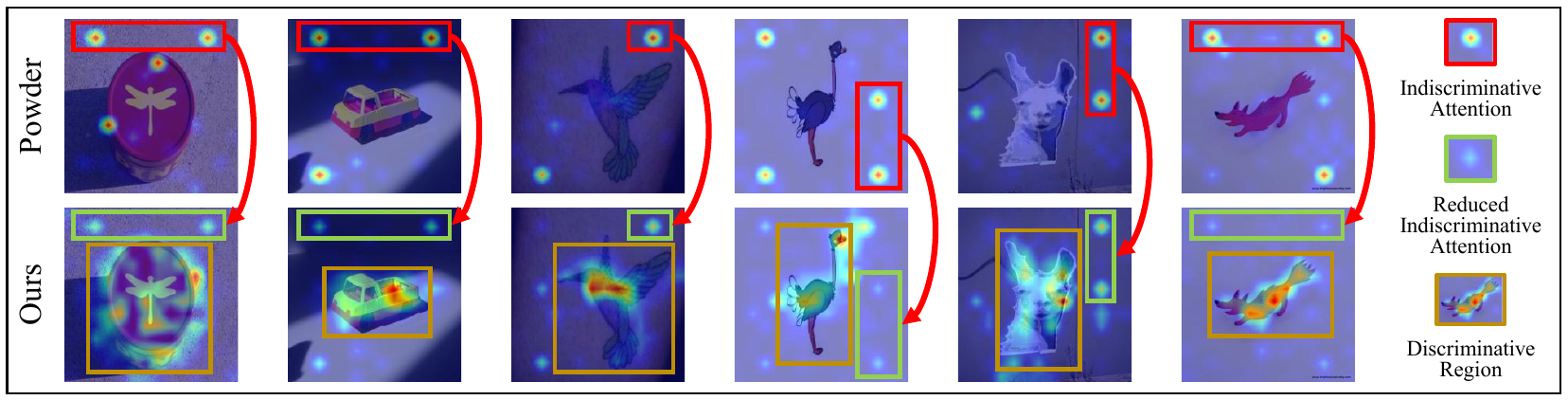} 
        \caption{\label{fig:heatmap-r} Visualization of attention across prompt and image regions.}   
\end{figure}

\begin{wrapfigure}{!t}{0.55\textwidth}
\includegraphics[width=0.55\textwidth]{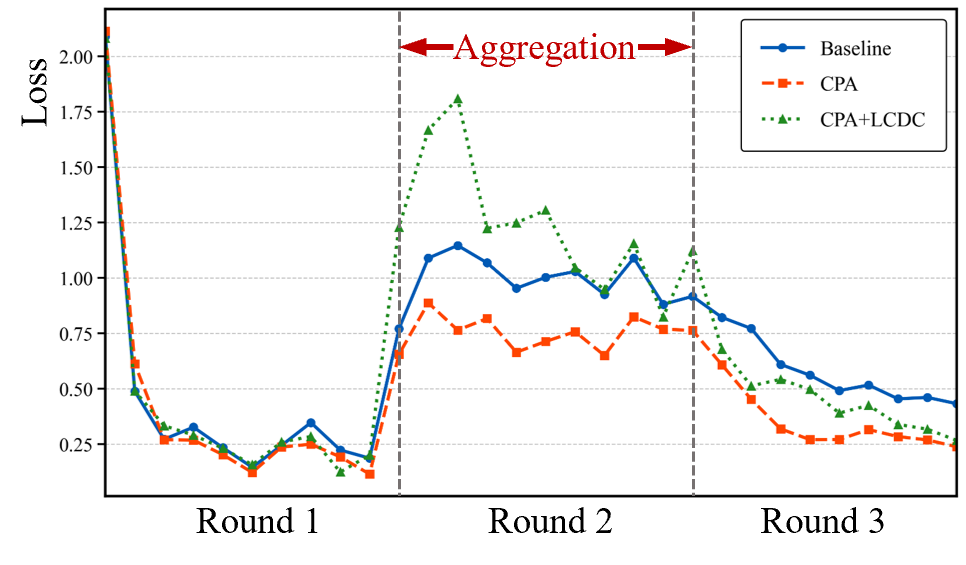}
\caption{Visualization of loss curves.}
\label{fig:loss}
\vspace{-0.4cm}
\end{wrapfigure}

\subsection{Ablation Study and Additional Analysis}

\textbf{Ablation on components}. 
Since Fig.~\ref{fig:framework} (a) and Fig.~\ref{fig:framework} (c) rely on each other, we present them as a unified component termed LCDC. Besides,  Fig.~\ref{fig:framework} (b) and Fig.~\ref{fig:framework} (d) also rely on each other, we present them as a unified component termed CPA. 

The ablation studies on LCDC and CPA are illustrated in Fig.~\ref{fig:ablation}, which are conducted with the ImageNet-R benchmark. When using LCDC module alone, our method obtains \textbf{1.88\%} improvement compared to the baseline, verifying the effectiveness of the Global Class Distribution Estimation and Local Class Distribution Compensation mechanism. Besides, CPA achieves \textbf{1.33\%} improvement compared to the baseline, demonstrating the effectiveness of our Class-aware Prompt Aggregation design. When all our modules are used together, the model performance is further improved with \textbf{2.51\%} improvement. This is because LCDC and CPA achieve input-level class information compensation and feature extraction parameter-level knowledge communication, respectively. Therefore, they are complementary to each other.

We also visualize the loss of Baseline, CPA and CPA+LCDC in Figure~\ref{fig:loss}. During round 1, different methods primarily learn with new data and converge similarly. When the first aggregation is conducted, all the methods show improved loss due to the parameter drift. CPA shows the least loss improvement due to the class-aware prompt fusion design that mitigates the knowledge conflict issue. CPA+LCDC shows a large loss improvement because the class distribution Compensation design guides the model in the early stage. During the second aggregation, the loss improvement of CPA+LCDC is significantly reduced since knowledge correlation between prompts is improved after round 2. After the training of round 3, both CPA and CPA+LCDC show significantly lower loss compared to the Baseline. Although CPA and CPA+LCDC obtain similar final losses, the performance of CPA+LCDC is superior to CPA since LCDC improves the robustness of the learned knowledge.

\textbf{Visualization of learned prompts:}
Figure~\ref{fig:heatmap-r} illustrates the prompt attention maps of our C${}^2$-Prompt in comparison with state-of-the-art Powder~\cite{piao2024federated}. 
Specifically, the prompts generated by Powder are largely dominated by class-irrelevant knowledge and exhibit limited discriminative feature extraction capacity. In contrast, the prompts generated by our method effectively focus on the discriminative regions and influence less on the class-irrelevant knowledge. These improvements are primarily attributed to our Class-aware Prompt Aggregation mechanism, which effectively alleviates conflicted knowledge fusion during prompt aggregation.

\begin{wraptable}{r}{0.55\linewidth}
\small
\centering
\caption{Comparison of communication and additional parameter overhead.}
\label{tab:overhead}
\resizebox{0.55\textwidth}{!}{
    \setlength\tabcolsep{2pt}
    \renewcommand\arraystretch{1}
\begin{tabular}{lccccc}
\toprule
        \multirow{2}{*}{Methods}& \multirow{2}{*}{Communication}& \multicolumn{2}{c}{Parameter}\\
         \cmidrule{3-4}
        &&Training&Inference\\
        \midrule
      Fed-L2P& 686.69MB& 3.96MB& 3.96MB\\
Fed-Dual &621.78MB& 4.73MB& 4.73MB\\
Fed-CODAP &815.63MB &11.43MB&11.43MB\\
Fed-CPrompt &815.63MB &11.43MB& 11.43M\\
Powder &\textbf{493.08MB}&\textbf{2.64MB}&\textbf{2.64MB}\\
\midrule
C${}^2$Prompt &\underline{496.01MB}& \underline{2.82MB}&\textbf{2.64MB}\\
        \bottomrule
\end{tabular}}
\label{components}
\end{wraptable}

\textbf{Communication Overhead:}
Table~\ref{tab:overhead} compares the communication and parameter overhead of C${}^2$-Prompt with state-of-the-art methods. Our approach demonstrates comparable communication and parameter costs with Powder, with only 0.6\% and 6.8\% increases, respectively. The additional communication overhead stems from the exchange of class distribution information between the server and clients. However, due to the sparse distribution of classes, this overhead remains minimal. The slight increase in training parameter count is attributed to the introduction of local class distribution compensation prompts, which are significantly fewer than the discriminative prompts commonly used in existing methods. Note that C${}^2$-Prompt \textbf{does not} introduce any additional parameters or computational overhead during inference, as only the discriminative prompts are employed for testing.

\section{Conclusion}
In this paper, we propose Class-aware Client Knowledge Interaction (C${}^2$Prompt), which enhances the class-wise knowledge coherence between prompts across clients, significantly alleviating both temporal and spatial forgetting by mitigating the potential knowledge conflict during prompt communication. C${}^2$Prompt introduces two kinds of prompts, local class distribution compensation prompt and local discriminativity prompt. The former transfers local class features to a global class-wise distribution to improve the intra-class semantic consistency across clients. The latter learn discrimination capacity with local data and aggregated with the ones from other clients in the server according to the class-wise affinity, enabling class-wise knowledge enhancement while alleviating conflicts. Extensive experiments on the challenging FCL benchmarks demonstrate that our method significantly outperforms the state-of-the-art, validating the effectiveness of our approach.

\textbf{Limitation Discussion:} Our approach requires class distribution communication at the initial round. Although this operation incurs minimal overhead due to the sparsity of distributional parameters, it introduces a minor communication cost. Furthermore, the prompt aggregation process generates client-specific prompts, slightly increasing computing and storage overhead compared to existing prompt-based methods. Nevertheless, this remains significantly more efficient than full fine-tuning approaches, as the number of learnable parameters in prompts is substantially smaller than that of the entire feature extractor.

\section*{Acknowledgement}

This work was supported by the National Key R\&D Program of China (2024YFA1410000), the National Natural Science Foundation of China (62376011), and the China Postdoctoral Science Foundation (2025T180417).






\bibliographystyle{IEEEtran}
\bibliography{neurips}

@String(PAMI = {IEEE Trans. Pattern Anal. Mach. Intell.})

@String(IJCV = {Int. J. Comput. Vis.})

@String(CVPR= {IEEE Conf. Comput. Vis. Pattern Recog.})

@String(ICCV= {Int. Conf. Comput. Vis.})

@String(ECCV= {Eur. Conf. Comput. Vis.})

@String(NeurIPS= {Adv. Neural Inform. Process. Syst.})

@String(ACMMM= {ACM Int. Conf. Multimedia})

@String(ICLR = {Int. Conf. Learn. Represent.})

@String(IJCAI = {IJCAI})

@String(AAAI = {AAAI})

@String(PAMI  = {IEEE TPAMI})

@String(IJCV  = {IJCV})

@String(CVPR  = {CVPR})

@String(ICCV  = {ICCV})

@String(ECCV  = {ECCV})

@String(NeurIPS= {NeurIPS})

@String(ACMMM = {ACM MM})

@String(ICLR  = {ICLR})

@inproceedings{wang2022learning,
  title={Learning to prompt for continual learning},
  author={Wang, Zifeng and Zhang, Zizhao and Lee, Chen-Yu and Zhang, Han and Sun, Ruoxi and Ren, Xiaoqi and Su, Guolong and Perot, Vincent and Dy, Jennifer and Pfister, Tomas},
  booktitle=CVPR,
  pages={139--149},
  year={2022}
}

@inproceedings{wang2022dualprompt,
  title={Dualprompt: Complementary prompting for rehearsal-free continual learning},
  author={Wang, Zifeng and Zhang, Zizhao and Ebrahimi, Sayna and Sun, Ruoxi and Zhang, Han and Lee, Chen-Yu and Ren, Xiaoqi and Su, Guolong and Perot, Vincent and Dy, Jennifer and others},
  booktitle=ECCV,
  pages={631--648},
  year={2022},
  organization={Springer}
}

@inproceedings{smith2023coda,
  title={Coda-prompt: Continual decomposed attention-based prompting for rehearsal-free continual learning},
  author={Smith, James Seale and Karlinsky, Leonid and Gutta, Vyshnavi and Cascante-Bonilla, Paola and Kim, Donghyun and Arbelle, Assaf and Panda, Rameswar and Feris, Rogerio and Kira, Zsolt},
  booktitle=CVPR,
  pages={11909--11919},
  year={2023}
}

@article{liu2024compositional,
  title={Compositional Prompting for Anti-Forgetting in Domain Incremental Learning},
  author={Liu, Zichen and Peng, Yuxin and Zhou, Jiahuan},
  journal={IJCV},
  pages={1--18},
  year={2024},
  publisher={Springer}
}

@inproceedings{xu2024distribution,
  title={Distribution-aware Knowledge Prototyping for Non-exemplar Lifelong Person Re-identification},
  author={Xu, Kunlun and Zou, Xu and Peng, Yuxin and Zhou, Jiahuan},
  booktitle=CVPR,
  pages={16604--16613},
  year={2024}
}

@inproceedings{li2024fcs,
  title={FCS: Feature Calibration and Separation for Non-Exemplar Class Incremental Learning},
  author={Li, Qiwei and Peng, Yuxin and Zhou, Jiahuan},
  booktitle=CVPR,
  pages={28495--28504},
  year={2024}
}

@inproceedings{xu2024lstkc,
  title={LSTKC: Long Short-Term Knowledge Consolidation for Lifelong Person Re-identification},
  author={Xu, Kunlun and Zou, Xu and Zhou, Jiahuan},
  booktitle=AAAI,
  volume={38},
  number={14},
  pages={16202--16210},
  year={2024}
}

@article{wang2024hierarchical,
  title={Hierarchical decomposition of prompt-based continual learning: Rethinking obscured sub-optimality},
  author={Wang, Liyuan and Xie, Jingyi and Zhang, Xingxing and Huang, Mingyi and Su, Hang and Zhu, Jun},
  journal=NeurIPS,
  volume={36},
  year={2023}
}

@article{li2024exemplar,
  title={Exemplar-Free Lifelong Person Re-identification via Prompt-Guided Adaptive Knowledge Consolidation},
  author={Li, Qiwei and Xu, Kunlun and Peng, Yuxin and Zhou, Jiahuan},
  journal={IJCV},
  pages={1--16},
  year={2024},
  publisher={Springer}
}

@inproceedings{bai2024diprompt,
  title={Diprompt: Disentangled prompt tuning for multiple latent domain generalization in federated learning},
  author={Bai, Sikai and Zhang, Jie and Guo, Song and Li, Shuaicheng and Guo, Jingcai and Hou, Jun and Han, Tao and Lu, Xiaocheng},
  booktitle=CVPR,
  pages={27284--27293},
  year={2024}
}

@inproceedings{feng2023learning,
  title={Learning federated visual prompt in null space for mri reconstruction},
  author={Feng, Chun-Mei and Li, Bangjun and Xu, Xinxing and Liu, Yong and Fu, Huazhu and Zuo, Wangmeng},
  booktitle=CVPR,
  pages={8064--8073},
  year={2023}
}

@article{hsu2019measuring,
  title={Measuring the effects of non-identical data distribution for federated visual classification},
  author={Hsu, Tzu-Ming Harry and Qi, Hang and Brown, Matthew},
  journal={arXiv preprint arXiv:1909.06335},
  year={2019}
}

@article{wang2020tackling,
  title={Tackling the objective inconsistency problem in heterogeneous federated optimization},
  author={Wang, Jianyu and Liu, Qinghua and Liang, Hao and Joshi, Gauri and Poor, H Vincent},
  journal=NeurIPS,
  volume={33},
  pages={7611--7623},
  year={2020}
}

@inproceedings{li2021fedbn,
  title={FedBN: Federated Learning on Non-IID Features via Local Batch Normalization},
  author={Li, Xiaoxiao and JIANG, Meirui and Zhang, Xiaofei and Kamp, Michael and Dou, Qi},
  booktitle={ICLR},
year={2021}
}

@article{li2020federated,
  title={Federated optimization in heterogeneous networks},
  author={Li, Tian and Sahu, Anit Kumar and Zaheer, Manzil and Sanjabi, Maziar and Talwalkar, Ameet and Smith, Virginia},
  journal={Proceedings of Machine learning and systems},
  volume={2},
  pages={429--450},
  year={2020}
}

@inproceedings{karimireddy2020scaffold,
  title={Scaffold: Stochastic controlled averaging for federated learning},
  author={Karimireddy, Sai Praneeth and Kale, Satyen and Mohri, Mehryar and Reddi, Sashank and Stich, Sebastian and Suresh, Ananda Theertha},
  booktitle={ICML},
  pages={5132--5143},
  year={2020},
  organization={PMLR}
}

@article{acar2021federated,
  title={Federated learning based on dynamic regularization},
  author={Acar, Durmus Alp Emre and Zhao, Yue and Navarro, Ramon Matas and Mattina, Matthew and Whatmough, Paul N and Saligrama, Venkatesh},
  journal={arXiv preprint arXiv:2111.04263},
  year={2021}
}

@inproceedings{gao2022feddc,
  title={Feddc: Federated learning with non-iid data via local drift decoupling and correction},
  author={Gao, Liang and Fu, Huazhu and Li, Li and Chen, Yingwen and Xu, Ming and Xu, Cheng-Zhong},
  booktitle=CVPR,
  pages={10112--10121},
  year={2022}
}

@article{lin2020ensemble,
  title={Ensemble distillation for robust model fusion in federated learning},
  author={Lin, Tao and Kong, Lingjing and Stich, Sebastian U and Jaggi, Martin},
  journal=NeurIPS,
  volume={33},
  pages={2351--2363},
  year={2020}
}

@inproceedings{zhu2021data,
  title={Data-free knowledge distillation for heterogeneous federated learning},
  author={Zhu, Zhuangdi and Hong, Junyuan and Zhou, Jiayu},
  booktitle={ICML},
  pages={12878--12889},
  year={2021},
  organization={PMLR}
}

@inproceedings{li2021model,
  title={Model-contrastive federated learning},
  author={Li, Qinbin and He, Bingsheng and Song, Dawn},
  booktitle=CVPR,
  pages={10713--10722},
  year={2021}
}

@inproceedings{guo2023pfedprompt,
  title={Pfedprompt: Learning personalized prompt for vision-language models in federated learning},
  author={Guo, Tao and Guo, Song and Wang, Junxiao},
  booktitle={Proceedings of the ACM Web Conference 2023},
  pages={1364--1374},
  year={2023}
}

@inproceedings{li2024global,
  title={Global and local prompts cooperation via optimal transport for federated learning},
  author={Li, Hongxia and Huang, Wei and Wang, Jingya and Shi, Ye},
  booktitle=CVPR,
  pages={12151--12161},
  year={2024}
}

@inproceedings{gao2024fedprok,
  title={Fedprok: Trustworthy federated class-incremental learning via prototypical feature knowledge transfer},
  author={Gao, Xin and Yang, Xin and Yu, Hao and Kang, Yan and Li, Tianrui},
  booktitle=CVPR,
  pages={4205--4214},
  year={2024}
}

@inproceedings{li2024personalized,
  title={Personalized federated domain-incremental learning based on adaptive knowledge matching},
  author={Li, Yichen and Xu, Wenchao and Wang, Haozhao and Qi, Yining and Guo, Jingcai and Li, Ruixuan},
  booktitle=ECCV,
  pages={127--144},
  year={2024},
  organization={Springer}
}

@inproceedings{tran2024text,
  title={Text-enhanced data-free approach for federated class-incremental learning},
  author={Tran, Minh-Tuan and Le, Trung and Le, Xuan-May and Harandi, Mehrtash and Phung, Dinh},
  booktitle=CVPR,
  pages={23870--23880},
  year={2024}
}

@inproceedings{zhang2021federated,
  title={Federated learning for internet of things},
  author={Zhang, Tuo and He, Chaoyang and Ma, Tianhao and Gao, Lei and Ma, Mark and Avestimehr, Salman},
  booktitle={Proceedings of the ACM conference on embedded networked sensor systems},
  pages={413--419},
  year={2021}
}

@inproceedings{li2024towards,
  title={Towards efficient replay in federated incremental learning},
  author={Li, Yichen and Li, Qunwei and Wang, Haozhao and Li, Ruixuan and Zhong, Wenliang and Zhang, Guannan},
  booktitle=CVPR,
  pages={12820--12829},
  year={2024}
}

@inproceedings{dong2022federated,
  title={Federated class-incremental learning},
  author={Dong, Jiahua and Wang, Lixu and Fang, Zhen and Sun, Gan and Xu, Shichao and Wang, Xiao and Zhu, Qi},
  booktitle=CVPR,
  pages={10164--10173},
  year={2022}
}

@inproceedings{shi2024clip,
  title={Clip-guided federated learning on heterogeneity and long-tailed data},
  author={Shi, Jiangming and Zheng, Shanshan and Yin, Xiangbo and Lu, Yang and Xie, Yuan and Qu, Yanyun},
  booktitle=AAAI,
  volume={38},
  number={13},
  pages={14955--14963},
  year={2024}
}

@inproceedings{huang2022learn,
  title={Learn from others and be yourself in heterogeneous federated learning},
  author={Huang, Wenke and Ye, Mang and Du, Bo},
  booktitle=CVPR,
  pages={10143--10153},
  year={2022}
}

@article{babakniya2023data,
  title={A data-free approach to mitigate catastrophic forgetting in federated class incremental learning for vision tasks},
  author={Babakniya, Sara and Fabian, Zalan and He, Chaoyang and Soltanolkotabi, Mahdi and Avestimehr, Salman},
  journal=NeurIPS,
  volume={36},
  pages={66408--66425},
  year={2023}
}

@inproceedings{guo2024pilora,
  title={Pilora: Prototype guided incremental lora for federated class-incremental learning},
  author={Guo, Haiyang and Zhu, Fei and Liu, Wenzhuo and Zhang, Xu-Yao and Liu, Cheng-Lin},
  booktitle=ECCV,
  pages={141--159},
  year={2024},
  organization={Springer}
}

@inproceedings{ma2025federated,
  title={Federated Few-Shot Class-Incremental Learning},
  author={Ma'sum, Muhammad Anwar and Pratama, Mahardhika and Liu, Lin and Habibullah, Habibullah and Kowalczyk, Ryszard},
  booktitle={ICLR},
  year={2025}
}

@inproceedings{piao2024federated,
  title={Federated continual learning via prompt-based dual knowledge transfer},
  author={Piao, Hongming and Wu, Yichen and Wu, Dapeng and Wei, Ying},
  booktitle={ICML},
  year={2024}
}

@inproceedings{xu2025dask,
  title={DASK: Distribution Rehearsing via Adaptive Style Kernel Learning for Exemplar-Free Lifelong Person Re-Identification},
  author={Xu, Kunlun and Jiang, Chenghao and Xiong, Peixi and Peng, Yuxin and Zhou, Jiahuan},
  booktitle=AAAI,
  volume={39},
  number={9},
  pages={8915--8923},
  year={2025}
}

@inproceedings{liu2025stop,
  title={STOP: Integrated Spatial-Temporal Dynamic Prompting for Video Understanding},
  author={Liu, Zichen and Xu, Kunlun and Su, Bing and Zou, Xu and Peng, Yuxin and Zhou, Jiahuan},
  booktitle=CVPR,
  year={2025}
}

@inproceedings{zhang2025scap,
  title={SCAP: Transductive Test-Time Adaptation via Supportive Clique-based Attribute Prompting},
  author={Zhang, Chenyu and Xu, Kunlun and Liu, Zichen and Peng, Yuxin and Zhou, Jiahuan},
  booktitle=CVPR,
  year={2025}
}

@inproceedings{xu2024mitigate,
  title={Mitigate Catastrophic Remembering via Continual Knowledge Purification for Noisy Lifelong Person Re-Identification},
  author={Xu, Kunlun and Zhang, Haozhuo and Li, Yu and Peng, Yuxin and Zhou, Jiahuan},
  booktitle=ACMMM,
  pages={5790--5799},
  year={2024}
}

@inproceedings{yoon2021federated,
  title={Federated continual learning with weighted inter-client transfer},
  author={Yoon, Jaehong and Jeong, Wonyong and Lee, Giwoong and Yang, Eunho and Hwang, Sung Ju},
  booktitle={ICML},
  pages={12073--12086},
  year={2021},
  organization={PMLR}
}

@inproceedings{ma2022continual,
  title={Continual Federated Learning Based on Knowledge Distillation.},
  author={Ma, Yuhang and Xie, Zhongle and Wang, Jue and Chen, Ke and Shou, Lidan},
  booktitle={IJCAI},
  pages={2182--2188},
  year={2022}
}

@inproceedings{shenaj2023asynchronous,
  title={Asynchronous federated continual learning},
  author={Shenaj, Donald and Toldo, Marco and Rigon, Alberto and Zanuttigh, Pietro},
  booktitle=CVPR,
  pages={5055--5063},
  year={2023}
}

@article{krizhevsky2009learning,
  title={Learning multiple layers of features from tiny images},
  author={Krizhevsky, Alex and Hinton, Geoffrey and others},
  year={2009},
  publisher={Toronto, ON, Canada}
}

@article{mcmahan2016federated,
  title={Federated learning of deep networks using model averaging},
  author={McMahan, H Brendan and Moore, Eider and Ramage, Daniel and y Arcas, Blaise Ag{\"u}era},
  journal={arXiv preprint arXiv:1602.05629},
  volume={2},
  number={2},
  pages={15--18},
  year={2016}
}

@article{salami2024closed,
  title={Closed-form merging of parameter-efficient modules for Federated Continual Learning},
  author={Salami, Riccardo and Buzzega, Pietro and Mosconi, Matteo and Bonato, Jacopo and Sabetta, Luigi and Calderara, Simone},
  journal={arXiv preprint arXiv:2410.17961},
  year={2024}
}

@inproceedings{bagwe2023fedcprompt,
title={Fed-{CP}rompt: Contrastive Prompt for Rehearsal-Free Federated Continual Learning},
author={Gaurav Bagwe and Xiaoyong Yuan and Miao Pan and Lan Zhang},
booktitle={Federated Learning and Analytics in Practice: Algorithms, Systems, Applications, and Opportunities},
year={2023}
}

@inproceedings{sun2025mos,
  title={Mos: Model surgery for pre-trained model-based class-incremental learning},
  author={Sun, Hai-Long and Zhou, Da-Wei and Zhao, Hanbin and Gan, Le and Zhan, De-Chuan and Ye, Han-Jia},
  booktitle=AAAI,
  volume={39},
  number={19},
  pages={20699--20707},
  year={2025}
}

@inproceedings{douillard2020podnet,
  title={Podnet: Pooled outputs distillation for small-tasks incremental learning},
  author={Douillard, Arthur and Cord, Matthieu and Ollion, Charles and Robert, Thomas and Valle, Eduardo},
  booktitle=ECCV,
  pages={86--102},
  year={2020},
  organization={Springer}
}

@inproceedings{chaudhry2018riemannian,
  title={Riemannian walk for incremental learning: Understanding forgetting and intransigence},
  author={Chaudhry, Arslan and Dokania, Puneet K and Ajanthan, Thalaiyasingam and Torr, Philip HS},
  booktitle=ECCV,
  pages={532--547},
  year={2018}
}

@article{lopez2017gradient,
  title={Gradient episodic memory for continual learning},
  author={Lopez-Paz, David and Ranzato, Marc'Aurelio},
  journal=NeurIPS,
  volume={30},
  year={2017}
}

@inproceedings{liang2024diffusion,
  title={Diffusion-driven data replay: A novel approach to combat forgetting in federated class continual learning},
  author={Liang, Jinglin and Zhong, Jin and Gu, Hanlin and Lu, Zhongqi and Tang, Xingxing and Dai, Gang and Huang, Shuangping and Fan, Lixin and Yang, Qiang},
  booktitle=ECCV,
  pages={303--319},
  year={2024},
  organization={Springer}
}

@inproceedings{qi2023better,
title={Better Generative Replay for Continual Federated Learning},
author={Daiqing Qi and Handong Zhao and Sheng Li},
booktitle={ICLR},
year={2023}
}

@inproceedings{odena2017conditional,
  title={Conditional image synthesis with auxiliary classifier gans},
  author={Odena, Augustus and Olah, Christopher and Shlens, Jonathon},
  booktitle={ICML},
  pages={2642--2651},
  year={2017},
  organization={PMLR}
}

@inproceedings{zhang2023target,
  title={Target: Federated class-continual learning via exemplar-free distillation},
  author={Zhang, Jie and Chen, Chen and Zhuang, Weiming and Lyu, Lingjuan},
  booktitle=ICCV,
  pages={4782--4793},
  year={2023}
}

@inproceedings{yu2024personalized,
  title={Personalized federated continual learning via multi-granularity prompt},
  author={Yu, Hao and Yang, Xin and Gao, Xin and Kang, Yan and Wang, Hao and Zhang, Junbo and Li, Tianrui},
  booktitle={SIGKDD},
  pages={4023--4034},
  year={2024}
}

@inproceedings{halbe2023hepco,
  title={HePCo: Data-Free Heterogeneous Prompt Consolidation for Continual Federated Learning},
  author={Halbe, Shaunak and Smith, James Seale and Tian, Junjiao and Kira, Zsolt},
  booktitle=NeurIPS
}

@article{liao2024foogd,
  title={FOOGD: Federated Collaboration for Both Out-of-distribution Generalization and Detection},
  author={Liao, Xinting and Liu, Weiming and Zhou, Pengyang and Yu, Fengyuan and Xu, Jiahe and Wang, Jun and Wang, Wenjie and Chen, Chaochao and Zheng, Xiaolin},
  journal=NeurIPS,
  volume={37},
  pages={132908--132945},
  year={2024}
}

@inproceedings{chenclassifier,
  title={Classifier Clustering and Feature Alignment for Federated Learning under Distributed Concept Drift},
  author={Chen, Junbao and Xue, Jingfeng and Wang, Yong and Liu, Zhenyan and Huang, Lu},
  booktitle=NeurIPS
}

@article{pan2024federated,
  title={Federated learning from vision-language foundation models: Theoretical analysis and method},
  author={Pan, Bikang and Huang, Wei and Shi, Ye},
  journal=NeurIPS,
  volume={37},
  pages={30590--30623},
  year={2024}
}

@article{zhang2024fedgmkd,
  title={Fedgmkd: An efficient prototype federated learning framework through knowledge distillation and discrepancy-aware aggregation},
  author={Zhang, Jianqiao and Shan, Caifeng and Han, Jungong},
  journal=NeurIPS,
  volume={37},
  pages={118326--118356},
  year={2024}
}

@inproceedings{wufiarse,
  title={FIARSE: Model-Heterogeneous Federated Learning via Importance-Aware Submodel Extraction},
  author={Wu, Feijie and Wang, Xingchen and Wang, Yaqing and Liu, Tianci and Su, Lu and Gao, Jing},
  booktitle=NeurIPS
}

@article{allouah2024fine,
  title={Fine-Tuning Personalization in Federated Learning to Mitigate Adversarial Clients},
  author={Allouah, Youssef and El Mrini, Abdellah and Guerraoui, Rachid and Gupta, Nirupam and Pinot, Rafael},
  journal=NeurIPS,
  volume={37},
  pages={100816--100844},
  year={2024}
}

@inproceedings{NEURIPS2019_15825aee,
 author = {Aljundi, Rahaf and Belilovsky, Eugene and Tuytelaars, Tinne and Charlin, Laurent and Caccia, Massimo and Lin, Min and Page-Caccia, Lucas},
 booktitle = NeurIPS,
 publisher = {Curran Associates, Inc.},
 title = {Online Continual Learning with Maximal Interfered Retrieval},
 volume = {32},
 year = {2019}
}

@inproceedings{NEURIPS2019_e562cd9c,
 author = {Aljundi, Rahaf and Lin, Min and Goujaud, Baptiste and Bengio, Yoshua},
 booktitle = NeurIPS,
 publisher = {Curran Associates, Inc.},
 title = {Gradient based sample selection for online continual learning},
 volume = {32},
 year = {2019}
}

@INPROCEEDINGS{9577808,
  author={Bang, Jihwan and Kim, Heesu and Yoo, YoungJoon and Ha, Jung-Woo and Choi, Jonghyun},
  booktitle=CVPR, 
  title={Rainbow Memory: Continual Learning with a Memory of Diverse Samples}, 
  year={2021},
  pages={8214-8223},
  doi={10.1109/CVPR46437.2021.00812}}

@article{khan2024hydra,
  title={HYDRA-FL: Hybrid Knowledge Distillation for Robust and Accurate Federated Learning},
  author={Khan, Momin Ahmad and Chandio, Yasra and Anwar, Fatima},
  journal=NeurIPS,
  volume={37},
  pages={50469--50493},
  year={2024}
}

@article{zhang2024improving,
  title={Improving Generalization in Federated Learning with Model-Data Mutual Information Regularization: A Posterior Inference Approach},
  author={Zhang, Hao and Li, Chenglin and Kan, Nuowen and Zheng, Ziyang and Dai, Wenrui and Zou, Junni and Xiong, Hongkai},
  journal=NeurIPS,
  volume={37},
  pages={136646--136678},
  year={2024}
}

@article{mclaughlin2024personalized,
  title={Personalized Federated Learning via Feature Distribution Adaptation},
  author={Mclaughlin, Connor and Su, Lili},
  journal=NeurIPS,
  volume={37},
  pages={77038--77059},
  year={2024}
}

@article{m2024personalized,
  title={Personalized federated learning with mixture of models for adaptive prediction and model fine-tuning},
  author={M Ghari, Pouya and Shen, Yanning},
  journal=NeurIPS,
  volume={37},
  pages={92155--92183},
  year={2024}
}

@article{weng2024probabilistic,
  title={Probabilistic Federated Prompt-Tuning with Non-IID and Imbalanced Data},
  author={Weng, Pei-Yau and Hoang, Minh and Nguyen, Lam and Thai, My T and Weng, Lily and Hoang, Nghia},
  journal=NeurIPS,
  volume={37},
  pages={81933--81958},
  year={2024}
}

@inproceedings{mendieta2022local,
  title={Local learning matters: Rethinking data heterogeneity in federated learning},
  author={Mendieta, Matias and Yang, Taojiannan and Wang, Pu and Lee, Minwoo and Ding, Zhengming and Chen, Chen},
  booktitle=CVPR,
  pages={8397--8406},
  year={2022}
}

@inproceedings{tan2022fedproto,
  title={Fedproto: Federated prototype learning across heterogeneous clients},
  author={Tan, Yue and Long, Guodong and Liu, Lu and Zhou, Tianyi and Lu, Qinghua and Jiang, Jing and Zhang, Chengqi},
  booktitle=AAAI,
  volume={36},
  number={8},
  pages={8432--8440},
  year={2022}
}

@inproceedings{yoonfedmix,
  title={FedMix: Approximation of Mixup under Mean Augmented Federated Learning},
  author={Yoon, Tehrim and Shin, Sumin and Hwang, Sung Ju and Yang, Eunho},
  booktitle={ICLR}
}

@inproceedings{hu2022fedsynth,
  title={FedSynth: Gradient Compression via Synthetic Data in Federated Learning},
  author={Hu, Shengyuan and Goetz, Jack and Malik, Kshitiz and Zhan, Hongyuan and Liu, Zhe and Liu, Yue},
  booktitle=NeurIPS
}

@inproceedings{xiong2023feddm,
  title={Feddm: Iterative distribution matching for communication-efficient federated learning},
  author={Xiong, Yuanhao and Wang, Ruochen and Cheng, Minhao and Yu, Felix and Hsieh, Cho-Jui},
  booktitle=CVPR,
  pages={16323--16332},
  year={2023}
}

@inproceedings{hendrycks2021many,
  title={The many faces of robustness: A critical analysis of out-of-distribution generalization},
  author={Hendrycks, Dan and Basart, Steven and Mu, Norman and Kadavath, Saurav and Wang, Frank and Dorundo, Evan and Desai, Rahul and Zhu, Tyler and Parajuli, Samyak and Guo, Mike and others},
  booktitle=ICCV,
  pages={8340--8349},
  year={2021}
}

@inproceedings{peng2019moment,
  title={Moment matching for multi-source domain adaptation},
  author={Peng, Xingchao and Bai, Qinxun and Xia, Xide and Huang, Zijun and Saenko, Kate and Wang, Bo},
  booktitle=ICCV,
  pages={1406--1415},
  year={2019}
}

@article{li2024resource,
  title={Resource-Aware Federated Self-Supervised Learning with Global Class Representations},
  author={Li, Mingyi and Zhang, Xiao and Wang, Qi and Liu, Tengfei and Wu, Ruofan and Wang, Weiqiang and Zhuang, Fuzhen and Xiong, Hui and Yu, Dongxiao},
  journal=NeurIPS,
  volume={37},
  pages={10008--10035},
  year={2024}
}

@inproceedings{wangtaming,
  title={Taming Cross-Domain Representation Variance in Federated Prototype Learning with Heterogeneous Data Domains},
  author={Wang, Lei and Bian, Jieming and Zhang, Letian and Chen, Chen and Xu, Jie},
  booktitle=NeurIPS
}

@inproceedings{morafahtowards,
  title={Towards Diverse Device Heterogeneous Federated Learning via Task Arithmetic Knowledge Integration},
  author={Morafah, Mahdi and Kungurtsev, Vyacheslav and Chang, Hojin Matthew and Chen, Chen and Lin, Bill},
  booktitle=NeurIPS
}

@article{kirkpatrick2017overcoming,
  title={Overcoming catastrophic forgetting in neural networks},
  author={Kirkpatrick, James and Pascanu, Razvan and Rabinowitz, Neil and Veness, Joel and Desjardins, Guillaume and Rusu, Andrei A and Milan, Kieran and Quan, John and Ramalho, Tiago and Grabska-Barwinska, Agnieszka and others},
  journal={Proceedings of the national academy of sciences},
  volume={114},
  number={13},
  pages={3521--3526},
  year={2017},
  publisher={National Academy of Sciences}
}

@article{li2017learning,
  title={Learning without forgetting},
  author={Li, Zhizhong and Hoiem, Derek},
  journal=PAMI,
  volume={40},
  number={12},
  pages={2935--2947},
  year={2017},
  publisher={IEEE}
}

@article{matena2022merging,
  title={Merging models with fisher-weighted averaging},
  author={Matena, Michael S and Raffel, Colin A},
  journal=NeurIPS,
  volume={35},
  pages={17703--17716},
  year={2022}
}

@article{jin2022dataless,
  title={Dataless knowledge fusion by merging weights of language models},
  author={Jin, Xisen and Ren, Xiang and Preotiuc-Pietro, Daniel and Cheng, Pengxiang},
  journal={arXiv preprint arXiv:2212.09849},
  year={2022}
}

@article{luo2021no,
  title={No fear of heterogeneity: Classifier calibration for federated learning with non-iid data},
  author={Luo, Mi and Chen, Fei and Hu, Dapeng and Zhang, Yifan and Liang, Jian and Feng, Jiashi},
  journal=NeurIPS,
  volume={34},
  pages={5972--5984},
  year={2021}
}

@inproceedings{xu2025componential,
  title={Componential Prompt-Knowledge Alignment for Domain Incremental Learning},
  author={Xu, Kunlun and Zou, Xu and Hua, Gang and Zhou, Jiahuan},
  booktitle={ICML}, 
  year={2025}
}

@article{xu2025long,
  title={Long Short-Term Knowledge Decomposition and Consolidation for Lifelong Person Re-Identification},
  author={Xu, Kunlun and Liu, Zichen and Zou, Xu and Peng, Yuxin and Zhou, Jiahuan},
  journal={TPAMI},
  year={2025},
  publisher={IEEE}
}

@article{zhou2025distribution,
  title={Distribution-Aware Knowledge Aligning and Prototyping for Non-Exemplar Lifelong Person Re-Identification},
  author={Zhou, Jiahuan and Xu, Kunlun and Zhuo, Fan and Zou, Xu and Peng, Yuxin},
  journal=PAMI,
  year={2025},
  publisher={IEEE}
}

@inproceedings{qi2025cross,
  title={Cross-Silo Feature Space Alignment for Federated Learning on Clients with Imbalanced Data},
  author={Qi, Zhuang and Meng, Lei and Li, Zhaochuan and Hu, Han and Meng, Xiangxu},
  booktitle=AAAI,
  pages={19986-19994},
  year={2025}
}

@inproceedings{qi2023cross,
  title={Cross-silo prototypical calibration for federated learning with non-iid data},
  author={Qi, Zhuang and Meng, Lei and Chen, Zitan and Hu, Han and Lin, Hui and Meng, Xiangxu},
  booktitle=ACMMM,
  pages={3099--3107},
  year={2023}
}

@article{meng2024improving,
  title={Improving global generalization and local personalization for federated learning},
  author={Meng, Lei and Qi, Zhuang and Wu, Lei and Du, Xiaoyu and Li, Zhaochuan and Cui, Lizhen and Meng, Xiangxu},
  journal={TNNLS},
page={76--87},
volume={36},
  year={2024},
  publisher={IEEE}
}


\appendix
\newpage

\section{Theoretical Justification of Distribution Operations}\label{sec:Justification}
\subsection{Server-side Distribution Aggregation}\label{sec:Server-side}
Federated continual learning (FCL) involves decentralized clients collaboratively learning over sequential tasks. To enhance cross-client coherence, we estimate the global class distributions on the server by aggregating local class distributions from each client. Given local class distributions $\mathcal{N}(\mu_1, \sigma_1^2)$, $\mathcal{N}(\mu_2, \sigma_2^2)$, $\dots$, $\mathcal{N}(\mu_n, \sigma_n^2)$ with probability density functions $f_1(x), f_2(x), \dots, f_n(x)$, the global distribution is defined using the clients' sample frequencies $p_1, p_2, \dots, p_n$, satisfying $\sum_{i=1}^n p_i = 1$.

The global class mean is calculated as:
\begin{equation}
\mu = \int \left[ p_1 f_1(x) + p_2 f_2(x) + \dots + p_n f_n(x) \right] x \, dx = p_1 \mu_1 + p_2 \mu_2 + \dots + p_n \mu_n
\end{equation}
The global class variance is expressed as:
\begin{equation}
\sigma^2 = \int \left[ p_1 f_1(x) + p_2 f_2(x) + \dots + p_n f_n(x) \right] x^2 \, dx - \mu^2
\end{equation}
Then we have
\begin{equation}
\sigma^2 = p_1 \left( \sigma_1^2 + \mu_1^2 \right) + p_2 \left( \sigma_2^2 + \mu_2^2 \right) + \dots + p_n \left( \sigma_n^2 + \mu_n^2 \right) - \mu^2
\end{equation}
This aggregation provides a comprehensive global distribution for each class, continuously updated as new tasks arrive. These global statistics are then communicated back to clients to guide local prompt optimization, enhancing semantic consistency.

\subsection{Local Class Distribution Compensation Loss}\label{sec:Client-side}
Upon receiving global class distributions, each client optimizes local class distribution compensation prompts. The derivation of the loss function for the prompt is as follows:

 For class \( i \), the global class distribution is represented as $\mathcal{N}(\mu_g^i, \boldsymbol{\Sigma}_g^i)$. Specifically, for feature vector \( X \in \mathbb{R}^d \) of class \( i \), its probability density is:
\begin{equation}
p(X \mid \mu_i, \boldsymbol{\Sigma}_i) = 
\frac{1}{(2\pi)^{d/2} |\boldsymbol{\Sigma}_i|^{1/2}} 
\cdot e^{ -\frac{1}{2} (X - \mu_i)^\top \boldsymbol{\Sigma}_i^{-1} (X - \mu_i) }
\label{eq1}
\end{equation}

where \( \mu_i \in \mathbb{R}^d \) is the mean vector and \( \boldsymbol{\Sigma}_i \in \mathbb{R}^{d \times d} \) is a positive definite symmetric covariance matrix. When assuming independent feature dimensions, the covariance matrix reduces to diagonal form:

\begin{equation}
\boldsymbol{\Sigma}_i = \text{diag}(\sigma_{i1}^2, \sigma_{i2}^2, \dots, \sigma_{id}^2)
\label{eq2}
\end{equation}

with determinant and inverse matrix given by:

\begin{equation}
|\boldsymbol{\Sigma}_i| = \prod_{j=1}^d \sigma_{ij}^2 \quad \text{and} \quad \boldsymbol{\Sigma}_i^{-1} = \text{diag}\left(\frac{1}{\sigma_{i1}^2}, \frac{1}{\sigma_{i2}^2}, \dots, \frac{1}{\sigma_{id}^2}\right)
\label{eq3}
\end{equation}

According to \eqref{eq1}, the exponent expands to:

\begin{equation}
(X - \mu_i)^\top \boldsymbol{\Sigma}_i^{-1}(X - \mu_i) = \sum_{j=1}^d \frac{(X_j - \mu_{ij})^2}{\sigma_{ij}^2}
\label{eq4}
\end{equation}

Thus, the probability density function decomposes as:

\begin{equation}
p_i(X|\mu_i,\sigma_i^2) = \frac{1}{(2\pi)^{d/2} \left( \prod_{j=1}^d \sigma_{ij}^2 \right)^{1/2}} 
\cdot e^{ -\frac{1}{2} \sum\limits_{j=1}^d \frac{(X_j - \mu_{ij})^2}{\sigma_{ij}^2} }
\label{eq5}
\end{equation}

In federated learning, client-generated features \( f_{x,p} \) should align with the server's global class distribution \( \mathcal{N}(\mu_g^i, \boldsymbol{\Sigma}_g^i) \). The optimization objective becomes maximizing the log-likelihood:

\begin{equation}
\mathcal{L}_c=\log p(f_{x,p} | \mu_g^i, \boldsymbol{\Sigma}_g^i)
\label{eq6}
\end{equation}

Substituting \eqref{eq1} and expanding:

\begin{equation}
\log p(f_{x,p} | \mu_g^i, \boldsymbol{\Sigma}_g^i) = -\frac{d}{2}\log(2\pi) - \frac{1}{2}\log|\boldsymbol{\Sigma}_g^i| - \frac{1}{2}(f_{x,p} - \mu_g^i)^\top (\boldsymbol{\Sigma}_g^i)^{-1} (f_{x,p} - \mu_g^i)
\label{eq7}
\end{equation}

Minimizing the negative log-likelihood loss:

\begin{equation}
\mathcal{L}_c = -\log p(f_{x,p} | \mu_g^i, \boldsymbol{\Sigma}_g^i) = \frac{d}{2}\log(2\pi) + \frac{1}{2}\log|\boldsymbol{\Sigma}_g^i| + \frac{1}{2}(f_{x,p} - \mu_g^i)^\top (\boldsymbol{\Sigma}_g^i)^{-1} (f_{x,p} - \mu_g^i)
\label{eq8}
\end{equation}

Since the first two constant terms can be omitted for optimization, $\mathcal{L}_c$ can be simplified to:

\begin{equation}
\mathcal{L}_c = \frac{1}{2}(f_{x,p} - \mu_g^i)^\top (\boldsymbol{\Sigma}_g^i)^{-1} (f_{x,p} - \mu_g^i)
\label{eq9}
\end{equation}












This objective encourages local features to match the global class distribution, effectively reducing inter-client distribution gaps and enhancing semantic coherence during federated updates.

\subsection{ Theoretical Implications}
Overall, the proposed distribution operations theoretically guarantee smoother knowledge alignment across clients by optimizing class-wise distribution coherence. This process stabilizes global knowledge representations throughout continual learning, improving the robustness of the learned knowledge and improving the aggregation compatibility between prompts from different clients, effectively mitigating spatial and temporal forgetting.

\begin{algorithm}[t!]
   \caption{Local Class Distribution Compensation and Global Class Distribution Estimation}
   \label{alg:framework}
\begin{algorithmic}
   \STATE {\bfseries Input:} Stage $t$ data $\mathcal{D}^t=\{\mathcal{D}^t_k\}_{k=1}^{K}$ 
   \STATE 
   \STATE {\bfseries Output:} Global prompt pool $\mathbf{P}_g^{t}$
   \STATE   
   \STATE 
   \STATE Initialize $\mathcal{P}_{d}^{t,k*}=None$
   
   \FOR{each round $r = 0$ to $N_r$}
   \IF{round r = 0}
        \STATE
        \STATE \textit{\# Local Class Distribution Compensation} ($\boldsymbol{LCDC}$)
       \FOR{each client $k$}
          \STATE Compute local class statistics $(\mu_{k,i}^t,\sigma_{k,i}^t)$ for each class $i$
          \STATE Upload $(\mu_{k,i}^t,\sigma_{k,i}^t)$ to the server
       \ENDFOR
       \STATE
       \STATE \textit{\# Global Class Distribution Estimation}
       \STATE Estimate global class center
       $\mu_i^g=\sum_{k=1}^K\mu_{i,k}^tp_{k,i}^t$, Eq.~\ref{eq:global-mu}
       \STATE Estimate global class variance
       $(\sigma_i^g)^2=\sum_{k=1}^K\left((\mu_{i,k}^t)^2+(\sigma_{i,k}^t)^2\right)p_{k,i}^t-(\mu_i^g)^2$, Eq.~\ref{eq:global-sigma}
       \STATE Distribute the global distribution to the corresponding clients
        \STATE
       \STATE \textit{\# Back to} ($\boldsymbol{LCDC}$)
       \FOR{each client $k=1$ to $K$}
          \STATE Initialize local class distribution complension prompts $\mathcal{P}^{c}_{t,k} = \{p^c_i\}_{i=1}^{|C_t^k|}$
          \STATE For input $x$, obtain $f_{x,p} = f_\theta([\boldsymbol{h}_x, \mathbf{p}^c_{\boldsymbol{x}}, \text{[CLS]}])$, Eq.~\ref{eq:cosine}
          \STATE Update $\mathcal{P}^{c}_{t,k}$ using           $\mathcal{L}_c=-\frac{1}{2}(f_{x,p}-\boldsymbol{\mu}_i^g)^\top(\boldsymbol{\Sigma}_i^g)^{-1}(f_{x,p}-\boldsymbol{\mu}_i^g)$, Eq.~\ref{eq:gaussian-loss}
       \ENDFOR
       \STATE Froze prompt $\mathcal{P}^{c}_{t,k}$
    \ENDIF

    \STATE
    \STATE \textit{\# Local Discriminativity Learning } 
    \FOR{each client $k=1$ to $K$}
        \IF{$\mathcal{P}^{d*}_{t,k}\neq None$}
         \STATE Initialize local discriminativity prompts $\mathcal{P}^{d}_{t,k} = \{p^d_i\}_{i=1}^N$ with $\mathcal{P}^{d*}_{t,k}$
        \ENDIF
         \STATE For each input $x$, pbtain $\mathbf{p}_{\boldsymbol{x}}^d$, $\mathbf{p}_{\boldsymbol{x}}^c$ and $H_{\boldsymbol{x}}$        
         \STATE Update using $\mathcal{L}_{ce}=CE\big(\mathbf{W}_k\boldsymbol{f}_\theta([\boldsymbol{h}_{\boldsymbol{x}},\mathbf{p}_{\boldsymbol{x}}^c,\mathbf{p}_{\boldsymbol{x}}^d,cls]),y\big)$, Eq.~\ref{eq:ce-loss}
         \STATE Obtain client histogram $H_k^i=\{s_c^j\}_{j=1}^{|\mathcal{C}_{k}^t|}$ for prompt $p_d^i$, where $s_c^j=\sum_{n=1}^{|D_{t,k}|}[H_{\boldsymbol{x}_n}]_j$, Eq.~\ref{eq:client-histogram}
         Upload $\{H_k^i\}_{i=1}^N$ and $\mathcal{P}^{d}_{t,k}$ to the server
    \ENDFOR

   \STATE    
   \STATE \textit{\# Class-aware Prompt Aggregation} ($\boldsymbol{CPA}$)

      \STATE Server collects all client histograms to form $\mathbf{H}_g^t\in\mathbb{R}^{KN\times |\mathcal{C}_t|}$
       \STATE Server collects all client discriminativity prompts to form $\mathbf{P}_g^t$
      \STATE Compute inter-prompt attention: $W_g^t=\gamma(\mathbf{H}_g^t{\mathbf{H}_g^{t^\top}}/\tau)$, Eq.~\ref{eq:correlation}
      \STATE Update prompts: $\mathbf{P}_g^{t*}=W_g^t\mathbf{P}_g^t$, Eq.~\ref{eq:weighted-aggregation}
      \STATE Distribute $\mathcal{P}^{d}_{t,k*}$ to corresponding clients
   \ENDFOR
   \STATE Return $\mathbf{P}_g^{t}$

\end{algorithmic}
\end{algorithm}

\section{Algorithm of the proposed approach}~\label{sec:alg}
The overall process of our C${}^2$Prompt is shown in Algorithm~\ref{alg:framework}.

\section{Details of the datasets and evaluation metrics}\label{appendix:D}
\subsection{Datasets}
We use three image datasets commonly utilized in Federated Continual Learning (FCL) to evaluate our method: ImageNet-R, DomainNet, and CIFAR-100. ImageNet-R consists of 30,000 images from 200 categories, including challenging samples from ImageNet and newly collected samples with various styles. The dataset is divided into a training set with 24,000 images and a test set with 6,000 images, and 20\% of the training set is selected as a validation set for tuning model parameters. DomainNet is a large dataset containing 600,000 images and 345 categories, spanning six different domains. CIFAR-100 contains 50,000 training and 10,000 test-colored images for 100 classes, respectively.
\subsection{Configuration of Federated Continual Learning Benchmarks}
The benchmark configuration of this paper follows previous FCL method Powder~\cite{piao2024federated}. Based on transferability (tasks have class overlap and each task contains only a small portion of each class’s data) and asynchrony, for ImageNet-R, each task randomly selects 20 classes (20\% samples per class), distributed randomly across clients with varying round durations. For DomainNet, each task randomly selects 35 classes (2\% samples per class due to closeness to pre-trained distribution, others same as ImageNet-R). We control task overlap by randomly selecting classes with the least overlap to study FCL performance under different task correlations. Unlike the common Dirichlet distribution method in FL, we avoid it here because in FCL, class sets of different tasks vary greatly, making it hard to control similarity with it. The setup details for the CIFAR-100 dataset are the same as those for ImageNet-R. Simultaneously, we set five clients for training, each executing distinct tasks. Furthermore, we organize the training process into phases, each consisting of three communication rounds. At the beginning of each phase, 40\% of the clients are selected to initiate learning on new tasks. To ensure the fairness of the results, we keep the optimizer, learning rate, and local training epochs consistent with those of the Powder method when training the classification prompts. Our LCDC is trained using the Adam optimizer when new tasks arrive, and the trained prompt is only used for the training of the formal classification prompt and not for the testing phase. 
\subsection{Evaluation Metrics}
We evaluate the effectiveness of our method by adapting seven metrics, including the Average accuracy of all tasks (Avg), Average Incremental Accuracy (AIA)\cite{douillard2020podnet}, Forgetting Measure (FM)\cite{chaudhry2018riemannian}, Forward Transfer (FT)\cite{lopez2017gradient}, Backward Transfer (BT)\cite{lopez2017gradient}, Combined Transfer (CT), Final Average Accuracy (FAA).
\paragraph{Average accuracy of all tasks (Avg)}
This metric measures the average accuracy of the final model across all tasks, computed as  
$$
\text{Avg} = \frac{1}{|\mathcal{T}|} \sum_{\mathcal{T}_c^t \in \mathcal{T}} a_{c, \max(\mathcal{R})}^t
$$  
where $\mathcal{T}$ denotes the set of all tasks during the Federated Continual Learning (FCL) process, and $a_{c, \max(\mathcal{R})}^t$ denotes the final accuracy of task $\mathcal{T}_c^t$ (i.e., the accuracy on this task when training concludes).
\paragraph{Average Incremental Accuracy (AIA)}
This metric measures the average accuracy over the FCL process, computed as $$\text{AIA} = \frac{1}{|\mathcal{R}|} \sum_{r \in \mathcal{R}} \sum_{\mathcal{T}_c^t \in \mathcal{T}_r} a_{c,r}^t$$ where $\mathcal{R}$ denotes the set of rounds with task switch, $\mathcal{T}_r$ denotes the set of existing tasks at round $r$, and $a_{c,r}^t$ denotes the accuracy of $\mathcal{T}_c^t$ at round $r$.
\paragraph{Forgetting Measure (FM)}
Forgetting is measured by the difference between the highest historical accuracy and the current accuracy of a task. This metric quantifies the model’s memory stability by the average forgetting over the FCL process, computed as $$\text{FM} = \frac{1}{|\mathcal{R}|} \sum_{r \in \mathcal{R}} \left( \sum_{\mathcal{T}_c^t \in \mathcal{T}_r} a_{c,r}^t - \tilde{a}_{c,r}^t \right)$$ where $\tilde{a}_{c,r}$ denotes the max accuracy of $\mathcal{T}_c^t$ before round $r$.
\paragraph{Forward Transfer (FT)}
This metric assesses the model’s ability to transfer knowledge into a task, from both previously learned tasks and other currently learned tasks, computed as $$\text{FT} = \frac{1}{|\mathcal{T}|} \sum_{\mathcal{T}_c^t \in \mathcal{T}} \left( \acute{a}_{c}^t - \hat{a}_{c}^t \right)$$ where $\mathcal{T}$ denotes all tasks during the FCL process, $\acute{a}_{c}^t$ denotes the accuracy of $\mathcal{T}_c^t$ when it finished, and $\hat{a}_{c}^t$ denotes the accuracy of single-task training.
\paragraph{Backward Transfer (BT)}
This metric evaluates the model’s ability to transfer knowledge from new tasks back to previously learned tasks, computed as $$\text{BT} = \frac{1}{|\mathcal{T}|} \sum_{\mathcal{T}_c^t \in \mathcal{T}} \left( a_{c, \max(\mathcal{R})}^t - \acute{a}_{c}^t \right)$$ where $a_{c, \max(\mathcal{R})}^t$ denotes the final accuracy of $\mathcal{T}_c^t$.
\paragraph{Combined Transfer (CT)}
This metric is a combination of FT and BT, evaluating the amount of information that a task $\mathcal{T}_c^t$ acquires from other tasks. The other tasks can have any sequence relationship with task $\mathcal{T}_c^t$ in terms of temporal dimension. It is computed as $$\text{CT} = \frac{1}{|\mathcal{T}|} \sum_{\mathcal{T}_c^t \in \mathcal{T}} \left( a_{c, \max(\mathcal{R})}^t - \hat{a}_{c}^t \right)$$
\paragraph{Final Average Accuracy (FAA)}
FAA is a standard metric used in FCIL to measure knowledge retention and accumulation. Let $a^t$ denote the test accuracy on the $t$-th task after the final incremental step. FAA is defined as:
$$\mathrm{FAA}=\frac{1}{T}\sum_{t=1}^Ta^t$$
where $T$ is the total number of tasks. A higher FAA indicates better overall performance across all tasks and stronger continual learning ability.

\section{Results under Other Federated Incremental Learning Experimental Settings}
In addition to the FCL setting proposed by Powder (ICML 2024)\cite{piao2024federated}, which considers task overlaps over time, we evaluate our approach under the federated class incremental learning (FCIL) setting, as investigated by PILoRA (ECCV 2024)\cite{guo2024pilora} and LoRM (ICLR 2025)~\cite{salami2024closed}.

\textbf{FCIL Setting:}
The FCIL setting divides the learning process into 10 incremental tasks, where class distributions are disjoint across tasks. For each task, training data is distributed among 10 clients following a Dirichlet distribution with parameter $\beta \in \{0.5, 0.1, 0.05\}$ to simulate non-IID scenarios. A smaller $\beta$ value represents a stronger data imbalance among clients.

\textbf{Training Details:}
A ViT-B/16 backbone pre-trained on ImageNet-21K is adopted. Each communication round consists of 5 training epochs, with a total of 5 communication rounds. Data augmentation for the training set includes random horizontal flipping and normalization. For the test set, preprocessing involves resizing with bicubic interpolation to $256 \times 256$, followed by center cropping to $224 \times 224$ and normalization.

\textbf{Comparison Results:}
We compare our C${}^2$Prompt with state-of-the-art FCIL methods~\cite{kirkpatrick2017overcoming,li2017learning,wang2022learning,smith2023coda,matena2022merging,jin2022dataless,luo2021no,tan2022fedproto,zhang2023target,guo2024pilora,salami2024closed} and the FCL method Powder~\cite{piao2024federated} on the ImageNet-R benchmark. The Final Average Accuracy (FAA)~\cite{salami2024closed} results are presented in Table~\ref{tab:fcil-imagenet-r}. The results demonstrate that our C${}^2$Prompt surpasses the state-of-the-art FCIL method LoRM, achieving improvements of \textbf{2.75\%/5.91\%/2.29\%} at $\beta = 0.5/0.1/0.05$, respectively. Furthermore, compared to the state-of-the-art prompt-based FCL method Powder, our approach achieves \textbf{0.61\%/2.60\%/6.48\%} improvements at $\beta = 0.5/0.1/0.05$, respectively. These increasing advantages under lower $\beta$ values are attributed to the effective inter-client intra-class distribution knowledge compensation mechanism, which significantly enhances model acquisition capacity and mitigates inter-client knowledge conflicts. These findings, alongside the experiments reported in the main paper, validate the adaptability of our approach to diverse practical federated continual learning scenarios.

\begin{table}[ht]
\centering
\caption{Performance comparison on ImageNet-R with different $\beta$ values.}
\begin{tabular}{llccc}
\toprule
\multirow{2}{*}{Method} & \multirow{2}{*}{Publication}&\multicolumn{3}{c}{\textbf{ImageNet-R (FAA)}} \\
\cmidrule(lr){3-5}
& &$\beta=0.5$ & $\beta=0.1$ & $\beta=0.05$   \\
\midrule
EWC \cite{kirkpatrick2017overcoming} &\scriptsize{\textcolor{darkgray}{\textit{NAS 2017}}}  & 58.93  & 48.15& 43.68   \\
LwF \cite{li2017learning}&\scriptsize{\textcolor{darkgray}{\textit{PAMI 2017}}}   & 54.03  & 41.02 & 46.07\\
FisherAVG \cite{matena2022merging}&\scriptsize{\textcolor{darkgray}{\textit{NeurIPS 2022}}}   & 58.68  & 50.82 & 47.33  \\
RegMean \cite{jin2022dataless} &\scriptsize{\textcolor{darkgray}{\textit{ICLR 2023}}}   & 61.18  & 57.00  & 55.80 \\
CCVR \cite{luo2021no} &\scriptsize{\textcolor{darkgray}{\textit{NeurIPS 2021}}}    & 70.00  & 62.60& 60.38   \\
L2P \cite{wang2022learning}  &\scriptsize{\textcolor{darkgray}{\textit{CVPR 2022}}}   & 42.08  & 23.85  &16.98  \\
CODA-P \cite{smith2023coda}  &\scriptsize{\textcolor{darkgray}{\textit{CVPR 2023}}}   & 61.18  & 36.73  & 25.82 \\
FedProto \cite{tan2022fedproto}&\scriptsize{\textcolor{darkgray}{\textit{AAAI 2022}}}  & 58.52  & 47.30 & 52.93  \\
TARGET \cite{zhang2023target} &\scriptsize{\textcolor{darkgray}{\textit{ICCV 2023}}}   & 54.65  & 45.83 & 41.32  \\
PILoRA \cite{guo2024pilora} &\scriptsize{\textcolor{darkgray}{\textit{ECCV 2024}}}    & 53.67  & 51.62 & 49.37  \\
Powder \cite{piao2024federated} &\scriptsize{\textcolor{darkgray}{\textit{ICML 2024}}} &
\underline{74.62} &\underline{67.14} &62.26 \\
LoRM \cite{salami2024closed} &\scriptsize{\textcolor{darkgray}{\textit{ICLR 2025}}}    & 72.48 & 63.83 &\underline{66.45} \\
\midrule
C${}^2$Prompt&\scriptsize{\textcolor{darkgray}{\textit{This Paper}}}   &\textbf{75.23}&\textbf{69.74}&\textbf{68.74}\\
\bottomrule
\end{tabular}
\label{tab:fcil-imagenet-r}
\end{table}

\begin{table}[ht]
\centering
\caption{Performance comparison on CIFAR-100 with different $\beta$ values.}
\begin{tabular}{llccc}
\toprule
\multirow{2}{*}{Method} & \multirow{2}{*}{Publication}&\multicolumn{3}{c}{\textbf{CIFAR-100 (FAA)}} \\
\cmidrule(lr){3-5}
& &$\beta=0.5$ & $\beta=0.1$ & $\beta=0.05$   \\
\midrule
EWC \cite{kirkpatrick2017overcoming} &\scriptsize{\textcolor{darkgray}{\textit{NAS 2017}}}  & 78.46&72.42&64.51   \\
LwF \cite{li2017learning}&\scriptsize{\textcolor{darkgray}{\textit{PAMI 2017}}}   & 62.87&55.56&47.09\\
FisherAVG \cite{matena2022merging}&\scriptsize{\textcolor{darkgray}{\textit{NeurIPS 2022}}}   & 76.10&74.43&65.31  \\
RegMean \cite{jin2022dataless} &\scriptsize{\textcolor{darkgray}{\textit{ICLR 2023}}}   & 59.80&45.88&39.08 \\
CCVR \cite{luo2021no} &\scriptsize{\textcolor{darkgray}{\textit{NeurIPS 2021}}}    & 79.95&75.14&65.30  \\
L2P \cite{wang2022learning}  &\scriptsize{\textcolor{darkgray}{\textit{CVPR 2022}}}   & 83.88&61.54&55.00 \\
CODA-P \cite{smith2023coda}  &\scriptsize{\textcolor{darkgray}{\textit{CVPR 2023}}}   & 82.25&61.82&46.74 \\
FedProto \cite{tan2022fedproto}&\scriptsize{\textcolor{darkgray}{\textit{AAAI 2022}}}  & 75.79&70.02&60.55  \\
TARGET \cite{zhang2023target} &\scriptsize{\textcolor{darkgray}{\textit{ICCV 2023}}}   & 74.72&72.32&62.60  \\
PILoRA \cite{guo2024pilora} &\scriptsize{\textcolor{darkgray}{\textit{ECCV 2024}}}    & 76.48&75.81&74.80  \\
Powder \cite{piao2024federated} &\scriptsize{\textcolor{darkgray}{\textit{ICML 2024}}} &
\underline{87.46} &\underline{85.33} &82.03 \\
LoRM \cite{salami2024closed} &\scriptsize{\textcolor{darkgray}{\textit{ICLR 2025}}}    & 86.95&81.76 &\underline{82.76} \\
\midrule
C${}^2$Prompt&\scriptsize{\textcolor{darkgray}{\textit{This Paper}}}   &\textbf{89.93}&\textbf{87.67}&\textbf{83.25}\\
\bottomrule
\end{tabular}
\label{tab:fcil-cifar-100}
\end{table}

\begin{table*}[htbp]
  \centering
    \caption{\label{tab:fcl-cifar-100} Result comparison on the CIFAR-100 benchmark}\label{table3}  
    \setlength{\tabcolsep}{0.75mm}{
    \begin{tabular*}{\textwidth}{@{\extracolsep{\fill}}ll>{\columncolor{hight_light}}c>{\columncolor{hight_light}}ccccccc}

        \toprule
        Methods&Publication& Avg$\uparrow$ & AIA$\uparrow$ & FM$\downarrow$ & FT$\uparrow$ & BT$\uparrow$ & CT$\uparrow$\\
        \midrule
        FedWEIT   &\scriptsize{\textcolor{darkgray}{\textit{ICML 2021}}}&95.17&95.61&0.48&\underline{3.76}&-0.91&\underline{3.04} \\
        CFeD    &\scriptsize{\textcolor{darkgray}{\textit{IJCAI 2022}}}&73.87&79.06&2.07&-11.01&-4.71&-14.78\\
        GLFC       &\scriptsize{\textcolor{darkgray}{\textit{CVPR 2022}}}&95.35&\underline{95.92}&0.35&\textbf{5.51}&-0.54&\textbf{5.08} \\
        Fedspace &\scriptsize{\textcolor{darkgray}{\textit{CVPR 2023}}}&94.17&94.87&1.03&0.37&-2.46&-1.60  \\
        \hline
        Fed-L2P& \scriptsize{\textcolor{darkgray}{\textit{CVPR 2022}}} & 95.65&95.68&\underline{0.08}&0.89&0.08&0.95   \\
        Fed-Dual & \scriptsize{\textcolor{darkgray}{\textit{ECCV 2022}}} &95.35&95.08&0.27&0.70&-0.24&0.51         \\
        Fed-CODAP&\scriptsize{\textcolor{darkgray}{\textit{ CVPR 2023}}}&82.05&55.71&13.77&-30.25&-18.60&-45.13\\
        FedCPrompt & \scriptsize{\textcolor{darkgray}{\textit{ICML 2023}}} &94.22&94.04&\underline{0.08}&0.32&0.00&0.32 \\        
        Powder &  \scriptsize{\textcolor{darkgray}{\textit{ICML 2024}}}&\underline{95.78}&95.83&0.35&2.03&-0.36&1.74     \\
        PILoRA&  \scriptsize{\textcolor{darkgray}{\textit{ECCV 2024}}}&76.21&82.31&0.32&0.01&-0.42&-0.40\\
        Fed-MOS&  \scriptsize{\textcolor{darkgray}{\textit{AAAI 2025}}}&85.11&87.23&0.20&-0.31&-0.11&-0.45\\        
        LoRM &  \scriptsize{\textcolor{darkgray}{\textit{ICLR 2025}}}&77.42&80.11&0.74&0.07&\underline{0.20}&0.22\\
        \hline
       Ours&\scriptsize{\textcolor{darkgray}{\textit{This Paper}}} &\textbf{97.32}&\textbf{96.78}&\textbf{-0.05}& 2.57&\textbf{0.31}&2.82\\

        \bottomrule
    \end{tabular*}}
\end{table*}%

\section{Experimental comparison on Cifar100}

\textbf{Avg Comparison:} Only our C${}^2$Prompt outperforms the state-of-the-art Powder, achieving improvements of \textbf{1.54\%} on Cifar100. This finding highlights the advantage of our approach in long-term knowledge consolidation. This can be attributed to the substantial enhancement of new knowledge acquisition capability achieved through our local class distribution compensation and class-aware discriminativity prompt aggregation strategies. Additionally, the precise knowledge communication mechanism prevents the fusion of irrelevant prompts, which would otherwise produce invalid prompts that are not only semantically divergent from new prompts but also clash with historical prompts.

\textbf{AIA Comparison:} Our C${}^2$Prompt achieves improvements of \textbf{0.86\%} on Cifar100, confirming that our approach consistently outperforms existing methods across various training stages. These improvements are due to the local class distribution compensation and class-aware discriminativity prompt aggregation designs, which strengthen robust local knowledge acquisition and enhance distributed knowledge collection.

\textbf{FM Comparison:} Our C${}^2$Prompt shows a negative forgetting rate on the small-scale dataset Cifar100. This suggests that new tasks can facilitate the learning of historical tasks when training samples are limited. These results confirm the effective antiforgetting capability of our method.

\textbf{FT Comparison:} Our C${}^2$Prompt shows advanced forward-transfer capacity, outperforming existing methods on Cifar100, respectively. This can be primarily attributed to two key components: the Global Class Distribution Estimation and Local Class Distribution Compensation mechanisms. Specifically, the former effectively leverages asynchronously arriving data from the same class to generate reliable global distribution estimates, while the latter utilizes these estimated global distributions to implement data-level information compensation, thereby significantly enhancing the learning efficiency of subsequent data.

\textbf{BT Comparison:} Our C${}^2$Prompt consistently demonstrates positive backward transfer capability on the Cifar100 dataset. This arises from the fact that asynchronously arriving data allow subsequent tasks to enhance knowledge of previously seen classes. We observe that C²Prompt's backward-transfer results relatively outperform those of Fed-CODAP and Powder. This is because the Local Class Distribution Compensation and Class-aware Prompt Aggregation designs significantly boost the distributed data learning capability at each stage, thereby leaving less improvement space for seen tasks.

\textbf{CT Comparison:} In terms of the comprehensive performance of forward and backward transfer, our C${}^2$Prompt overall outperforms other existing methods that employ efficient fine-tuning. These results demonstrate that the class-aware client knowledge interaction designs proposed in this paper effectively enhance the overall learning capability of Federated Continual Learning (FCL) in the temporal dimension. Specifically: the Global Class Distribution Estimation module efficiently aggregates distributional information across spatial and temporal data sources; the Local Class Distribution Compensation module utilizes global distribution representations to overcome the non-IID (non-independent and identically distributed) phenomenon across clients; and the Local Discriminativity Learning and Class-aware Prompt Aggregation modules effectively integrate distributional knowledge into prompts.

\begin{figure}[!t]
\centering
\begin{subfigure}[t]{0.48\textwidth}
    \centering
    \includegraphics[width=\textwidth]{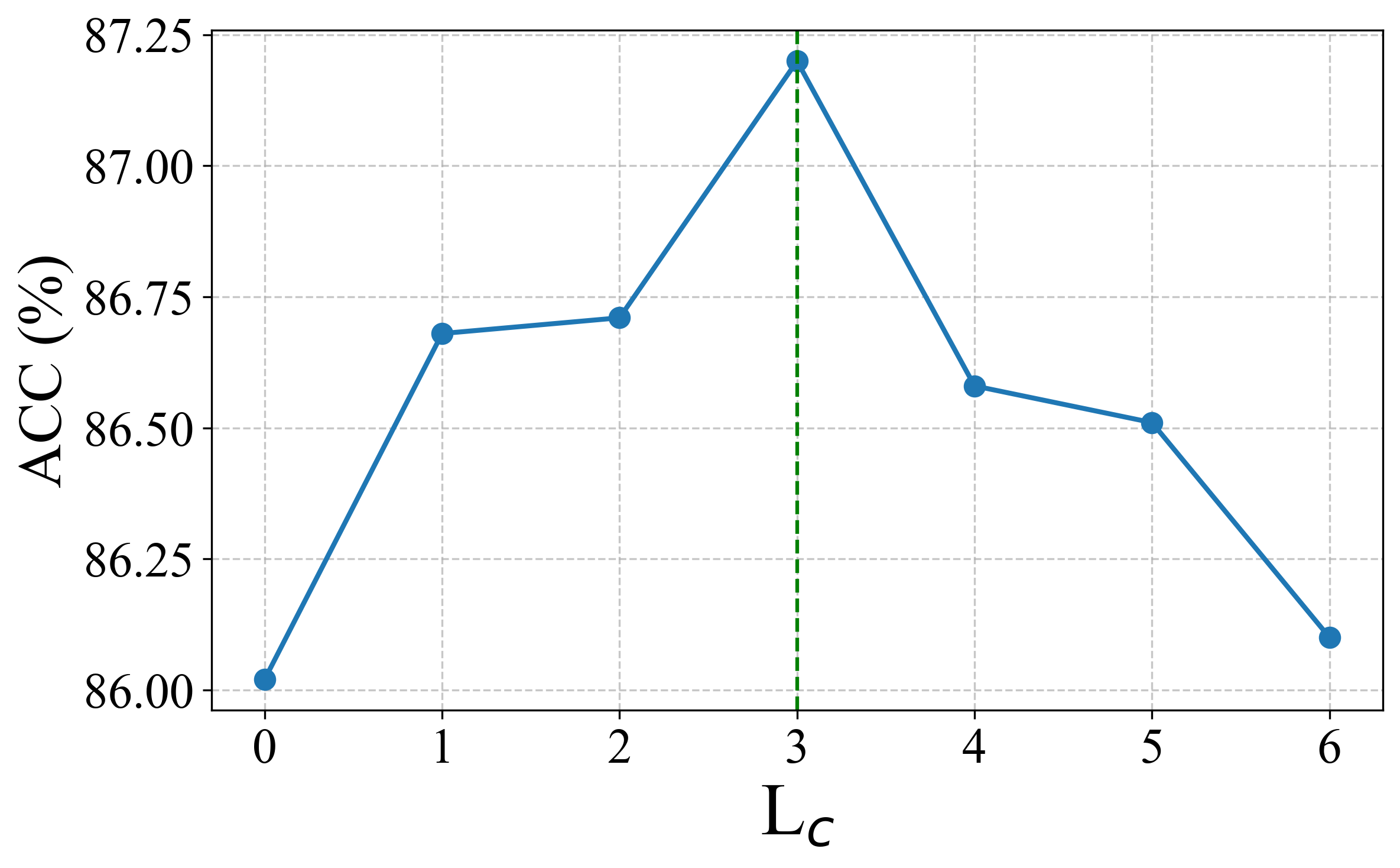}
\end{subfigure}
\hfill
\begin{subfigure}[t]{0.48\textwidth}
    \centering
    \includegraphics[width=\textwidth]{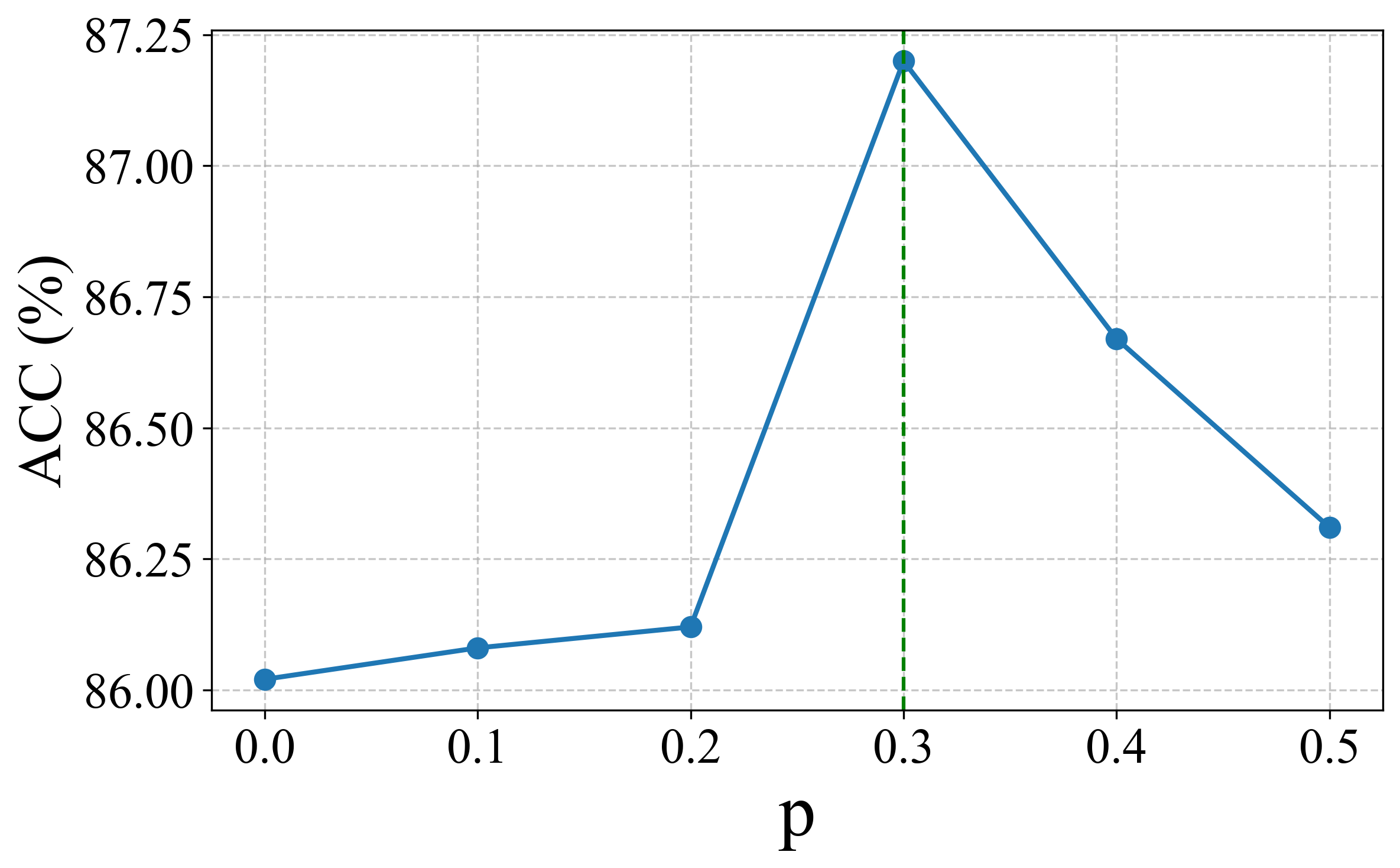}
\end{subfigure}
\caption{Ablation studies on the hyper-parameters under
ImageNet-R dataset.}
\label{fig:hyper-parameters}
\end{figure}
\begin{figure}[t]
    \centering
	\includegraphics[width=1.0\linewidth]{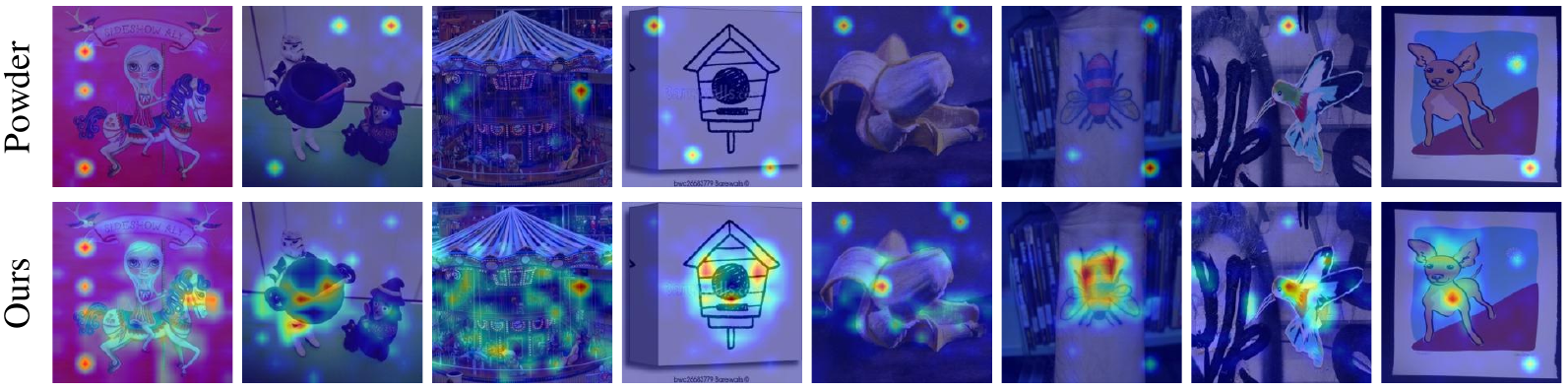} 
        \caption{\label{fig:heatmap-r-2} Prompt attention visualization on ImageNet-R.}   
\end{figure}
\begin{figure}[t]
    \centering
	\includegraphics[width=1.0\linewidth]{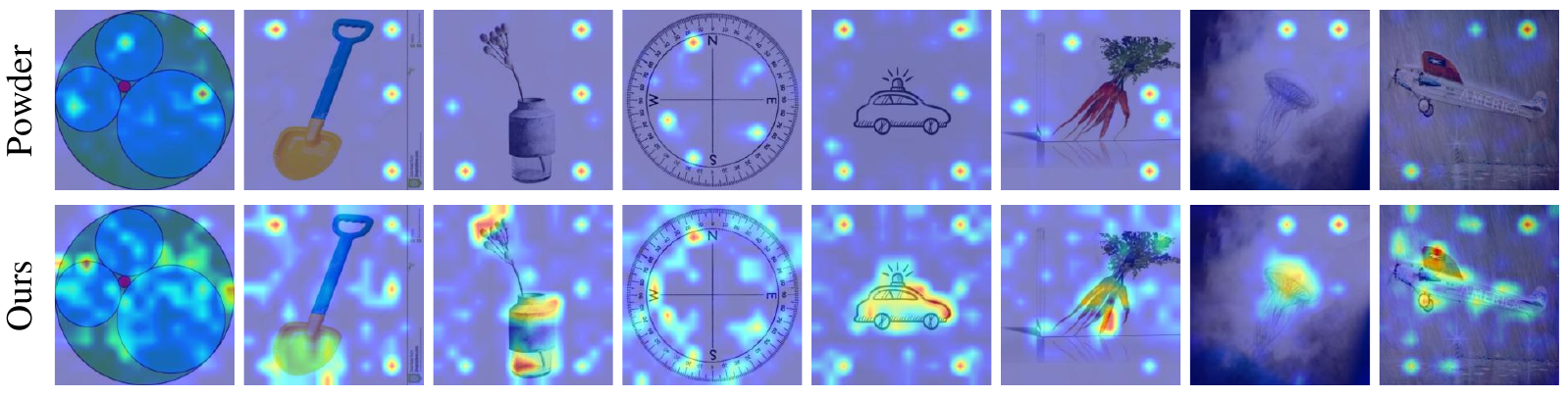} 
        \caption{\label{fig:heatmap-d} Prompt attention visualization on DomainNet.}   
\end{figure}

\section{Analysis on the hyper-parameters}
In Figure~\ref{fig:hyper-parameters}, we evaluate the performance of C${}^2$Prompt under different values of the hyper-parameters $L_c$ and $p$. The parameter $L_c$ represents the length of the local class distribution compensation prompts. When the prompt length is less than or equal to 3, a larger value of parameter a enables the trained prompts to better fit the central distribution of the class, thereby improving the model's performance. Meanwhile, $p$ serves as the usage probability of local class distribution compensation prompts, is used to determine the number of generated new central distribution samples. Based on experimental analysis, the optimal hyperparameter value for $p$ is set to 0.3.

\section{Prompt attention visualization comparison on different benchmarks}

 Figure~\ref{fig:heatmap-r-2} and Figure~\ref{fig:heatmap-d} present the visualization comparison of prompt attention maps between our C${}^2$Prompt framework and the state-of-the-art Powder method across the challenging ImageNet-R and DomainNet benchmarks. The results demonstrate that prompts generated by Powder are predominantly influenced by class-irrelevant knowledge, leading to limited discriminative feature extraction capabilities. In contrast, our method effectively focuses prompt activations on discriminative regions while significantly reducing interference from class-agnostic knowledge. These improvements are primarily attributed to the proposed Class-aware Prompt Aggregation mechanism, which systematically alleviates the fusion of knowledge conflicts during prompt aggregation through explicit semantic alignment.

 \section{Broader Impacts}
Our method tackles a practical federated continual learning (FCL) problem and introduces a novel approach that effectively improves the local parameter learning in the client side and enhances knowledge aggregation capacity on the server side. 

\textbf{The Potential Positive Societal Impacts of this research include:}

\textit{1. Enhanced Privacy Preservation in Decentralized Learning}

Federated continual learning (FCL) inherently avoids centralized data collection. C$^{2}$Prompt further eliminates reliance on raw data or generative models for knowledge retention, reducing risks of sensitive data leakage. This is critical for applications like healthcare (e.g., personalized disease prediction across hospitals) or finance (e.g., fraud detection without sharing transaction details).

\textit{2. Improved Adaptability in Dynamic Environments}

By addressing both temporal and spatial forgetting, C$^{2}$Prompt enables models to continuously adapt to evolving data streams. This ensures long-term reliability in scenarios where data distributions shift over time or vary across regions.

\textit{3. Democratization of AI in Resource-Constrained Settings}

The lightweight prompt-based framework reduces computational and communication overhead compared to traditional methods. This democratizes access to AI for edge devices with limited resources (e.g., rural IoT sensors, low-power medical devices), fostering equitable technological progress.

\textit{4. Mitigation of Model Bias via Class-Aware Aggregation}

The class-aware prompt aggregation (CPA) mechanism explicitly accounts for inter-client class relevance, potentially reducing biases arising from skewed local data distributions. For instance, in facial recognition systems deployed across diverse demographics, CPA could improve fairness by ensuring minority groups’ features are adequately represented.

\textbf{The Potential Negative Societal Impacts of this research include:}

\textit{1. Energy Consumption}

Additional distributional information communication and client-wise aggregation across distributed clients may increase energy consumption, particularly in large-scale deployments.

\newpage
\section*{NeurIPS Paper Checklist}

\begin{enumerate}

\item {\bf Claims}
    \item[] Question: Do the main claims made in the abstract and introduction accurately reflect the paper's contributions and scope?
    \item[] Answer: \answerYes{} 
    \item[] Justification: The main claims made in the abstract and introduction accurately reflect the
 paper’s contributions and scope.

\item {\bf Limitations}
    \item[] Question: Does the paper discuss the limitations of the work performed by the authors?
    \item[] Answer: \answerYes{} 
    \item[] Justification: We have analyzed the limitations of this method over computing and communication overheads.

\item {\bf Theory assumptions and proofs}
    \item[] Question: For each theoretical result, does the paper provide the full set of assumptions and a complete (and correct) proof?
    \item[] Answer: \answerYes{} 
    \item[] Justification: We have provided a theoretical analysis of our distribution operation with a complete theoretical derivation proof.

    \item {\bf Experimental result reproducibility}
    \item[] Question: Does the paper fully disclose all the information needed to reproduce the main experimental results of the paper to the extent that it affects the main claims and/or conclusions of the paper (regardless of whether the code and data are provided or not)?
    \item[] Answer: \answerYes{} 
    \item[] Justification: The paper fully discloses all the information needed to reproduce the main experimental results of the paper to the extent that it affects the main claims and/or conclusions of the paper.

\item {\bf Open access to data and code}
    \item[] Question: Does the paper provide open access to the data and code, with sufficient instructions to faithfully reproduce the main experimental results, as described in supplemental material?
    \item[] Answer: \answerYes{} 
    \item[] Justification: We have attached our source code and data access links in the supplementary materials, with sufficient instructions to faithfully reproduce the main experimental results.

\item {\bf Experimental setting/details}
    \item[] Question: Does the paper specify all the training and test details (e.g., data splits, hyperparameters, how they were chosen, type of optimizer, etc.) necessary to understand the results?
    \item[] Answer: \answerYes{} 
    \item[] Justification: We have specified all the training and test details necessary to understand the results.

\item {\bf Experiment statistical significance}
    \item[] Question: Does the paper report error bars suitably and correctly defined or other appropriate information about the statistical significance of the experiments?
    \item[] Answer: \answerNo{} 
    \item[] Justification: Error bars are not reported due to computational constraints. However, all experiments are conducted with consistent benchmark configurations and fixed random seeds, ensuring reproducibility. The significant and consistent improvements across multiple benchmarks substantiate the effectiveness of our method.

\item {\bf Experiments compute resources}
    \item[] Question: For each experiment, does the paper provide sufficient information on the computer resources (type of compute workers, memory, time of execution) needed to reproduce the experiments?
    \item[] Answer: \answerYes{} 
    \item[] Justification: The paper provide sufficient information on the computer resources.
    
\item {\bf Code of ethics}
    \item[] Question: Does the research conducted in the paper conform, in every respect, with the NeurIPS Code of Ethics \url{https://neurips.cc/public/EthicsGuidelines}?
    \item[] Answer: \answerYes{} 
    \item[] Justification: The research conducted in the paper conform, in every respect, with the NeurIPS Code of Ethics.


\item {\bf Broader impacts}
    \item[] Question: Does the paper discuss both potential positive societal impacts and negative societal impacts of the work performed?
    \item[] Answer: \answerYes{} 
    \item[] Justification: The Broader impacts has been discussed in the Appendix.
    
\item {\bf Safeguards}
    \item[] Question: Does the paper describe safeguards that have been put in place for responsible release of data or models that have a high risk for misuse (e.g., pretrained language models, image generators, or scraped datasets)?
    \item[] Answer: \answerNA{} 
    \item[] Justification: The paper poses no such risks.

\item {\bf Licenses for existing assets}
    \item[] Question: Are the creators or original owners of assets (e.g., code, data, models), used in the paper, properly credited and are the license and terms of use explicitly mentioned and properly respected?
    \item[] Answer: \answerYes{} 
    \item[] Justification: They are properly credited and respected.

\item {\bf New assets}
    \item[] Question: Are new assets introduced in the paper well documented and is the documentation provided alongside the assets?
    \item[] Answer: \answerNA{} 
    \item[] Justification: The paper does not release new assets.

\item {\bf Crowdsourcing and research with human subjects}
    \item[] Question: For crowdsourcing experiments and research with human subjects, does the paper include the full text of instructions given to participants and screenshots, if applicable, as well as details about compensation (if any)? 
    \item[] Answer: \answerNA{} 
    \item[] Justification: This paper does not involve crowdsourcing nor research with human subjects.

\item {\bf Institutional review board (IRB) approvals or equivalent for research with human subjects}
    \item[] Question: Does the paper describe potential risks incurred by study participants, whether such risks were disclosed to the subjects, and whether Institutional Review Board (IRB) approvals (or an equivalent approval/review based on the requirements of your country or institution) were obtained?
    \item[] Answer: \answerNA{} 
    \item[] Justification:  This paper does not involve crowdsourcing nor research with human subjects.

\item {\bf Declaration of LLM usage}
    \item[] Question: Does the paper describe the usage of LLMs if it is an important, original, or non-standard component of the core methods in this research? Note that if the LLM is used only for writing, editing, or formatting purposes and does not impact the core methodology, scientific rigorousness, or originality of the research, declaration is not required.
    \item[] Answer: \answerNA{} 
    \item[] Justification: This core method development in this research does not involve LLMs as any important, original, or non-standard components.

\end{enumerate}

\end{document}